\documentclass{article}

\usepackage{microtype}
\usepackage{graphicx}
\usepackage{xcolor}
\usepackage{subcaption}
\usepackage{booktabs} 
\usepackage{multirow}
\usepackage{hyperref}
\usepackage{colortbl}

\usepackage[accepted]{icml2026}

\usepackage{amsmath}
\usepackage{amssymb}
\usepackage{mathtools}
\usepackage{amsthm}
\usepackage{adjustbox}
\usepackage{xspace}

\usepackage[capitalize,noabbrev]{cleveref}

\theoremstyle{plain}

\theoremstyle{definition}

\theoremstyle{remark}

\newcommand{\michi}[1]{\textcolor{cyan!90!black}{\\michi: #1}}
\newcommand{\hector}[1]{\textcolor{red!90!black}{\\hector: #1}}
\newcommand{\yannik}[1]{\textcolor{green!90!black}{\\yannik: #1}}

\def\multiset#1{\ensuremath{ \left\{\kern-.35em\left\{ #1 \right\}\kern-.35em\right\}}}

\definecolor{covred}{rgb}{0.81, 0.09, 0.13}
\definecolor{covgreen}{rgb}{0.13, 0.55, 0.13}
\definecolor{covbg}{RGB}{250,250,250}

\definecolor{sokcol}{rgb}{0.19, 0.55, 0.91}
\definecolor{witcol}{HTML}{2CA02C}  
\definecolor{tilecol}{rgb}{0.93, 0.53, 0.18}

\newcommand{\triple}[3]{%
\makebox[2.6em][c]{\textcolor{sokcol}{#1}} /
\makebox[2.6em][c]{\textcolor{witcol}{#2}} /
\makebox[2.6em][c]{\textcolor{tilecol}{#3}}}

\newcommand{\wtriple}[3]{%
\makebox[2.8em][c]{\textcolor{sokcol}{#1}} /
\makebox[2.9em][c]{\textcolor{witcol}{#2}} /
\makebox[2.6em][c]{\textcolor{tilecol}{#3}}}

\newcommand{\double}[2]{%
\makebox[1.em][r]{\textcolor{sokcol}{#1}}\hspace{1pt}/\hspace{1pt}%
\makebox[1.em][l]{\textcolor{witcol}{#2}}}

\newcommand{\wdouble}[2]{%
\makebox[1.em][r]{\textcolor{sokcol}{#1}}\hspace{1pt}/\hspace{1pt}%
\makebox[1.em][l]{\textcolor{witcol}{#2}}}

\newcommand{\sdouble}[2]{%
\makebox[2.5em][r]{\textcolor{sokcol}{#1}}\hspace{1pt}/\hspace{1pt}%
\makebox[2.5em][l]{\textcolor{witcol}{#2}}}

\newcommand{\swdouble}[2]{%
\makebox[2.5em][r]{\textcolor{sokcol}{#1}}\hspace{1pt}/\hspace{1pt}%
\makebox[2.5em][l]{\textcolor{witcol}{#2}}}

\newcommand{\planpair}[2]{%
  \makebox[4em][r]{#1}\,$/$\,\makebox[2.5em][l]{#2}%
}
\newcommand{\planpairl}[2]{%
  \makebox[2em][r]{#1}\,$/$\,\makebox[2.5em][l]{#2}%
}

\newcommand{\model}{\mathrm{GSP}}

\newcommand{\astar}{\text{A}\!^\star}
\newcommand{\wastar}{\text{WA}\!^\star}
\newcommand{\lrtastar}{\text{LRTA}\!^\star}

\newcommand{\pred}[1]{\emph{#1}}
\newcommand*{\eg}{e.g.\@\xspace}
\newcommand*{\ie}{i.e.\@\xspace}

\newcommand{\Omit}[1]{}

\usepackage[textsize=tiny]{todonotes}

\icmltitlerunning{Learning to Search and Searching to Learn for Generalization in Planning}
\begin{document}

\twocolumn[
  \icmltitle{Learning to Search and Searching to Learn for Generalization in Planning}



  \icmlsetsymbol{equal}{*}

  \begin{icmlauthorlist}
    \icmlauthor{Michael Aichmüller*}{rwth}
    \icmlauthor{Yannik Hesse*}{rwth}
    \icmlauthor{Hector Geffner}{rwth}
  \end{icmlauthorlist}

  \icmlaffiliation{rwth}{Department of Machine Learning and Reasoning, RWTH Aachen University, Aachen, Germany}

  \icmlcorrespondingauthor{Michael Aichmüller}{michael.aichmueller@ml.rwth-aachen.de}

  \icmlkeywords{Machine Learning, ICML, AI, Reinforcement Learning, Classical Planning, Graph Neural Networks}

  \vskip 0.3in
]



\printAffiliationsAndNotice{\icmlEqualContribution}

\begin{abstract}
Combinatorial generalization remains a central challenge in Deep Reinforcement Learning (DRL). Classical planning provides a simple yet challenging setting to study this problem through explicit relational descriptions, without requiring learning from perception. 
In sparse-reward domains, standard RL exploration via real-time search is ineffective, and learning-based planning methods often rely on expert demonstrations, hindsight relabeling, or random walks from the goal state. In contrast, planners rely on best-first search methods such as $\astar{}$ to solve problems from scratch. We propose a self-improving $\wastar{}$ learning framework in combination with a value heuristic represented by a Relational Graph Neural Network: the heuristic guides search, and the resulting search data updates the heuristic via $Q$-learning. This loop yields heuristics that can function as general policies and solve new instances even without search, where DRL otherwise fails, as we show on puzzles such as Sokoban, PushWorld, The Witness, and the 2023 International Planning Competition benchmarks. Notably, we demonstrate strong zero-shot generalization: For example, heuristics trained on Blocksworld instances with fewer than 30 blocks successfully solve instances with 488 blocks without search.
\end{abstract}

\section{Introduction}

Combinatorial  generalization is a key challenge in deep reinforcement learning where the learned policies or value functions
are expected to generalize out-of-distribution due to a common problem structure \cite{kirk2023survey,lake2023human,mohan2024structure}.
 Classical planning is an ideal setting for studying and addressing this problem because the common problem structure is given and  does not need to be learned
from pixels \cite{russell:book,ghallab:book,geffner:book}. 
\Omit{
Classical planning problems are indeed  deterministic goal-reaching MDPs expressed in relational languages that
make a clean separation between a planning domain (e.g., Blocks or Sokoban), and the domain instances, which 
differ from   each other in the number of objects, initial state, and goal. 
The states $s$ of all classical planning problems are given by sets  of atoms of the form $p(o_1, \ldots, o_k)$, where 
$p$ is a domain predicate or relation, $k$ is the predicate arity,
and $o_1$ to $o_k$ are objects (logically, constant symbols). The different instances of a given domain do not share the
the objects but they all share the same finite set of predicates or relations $p$. For instance, the states of any   Blocksworld instance
can be expressed in terms of the fixed   set of predicates  $on(x,y)$, $hold(x)$, \emph{hand-free}, and $ontable(x)$.}
A planning \emph{domain} specifies a fixed relational vocabulary and action schemas, while \emph{instances} vary in the number of objects, the initial state, and the goal. This clean domain--instance separation induces systematic out-of-distribution shifts, including changes in branching factor and required search depth, and makes classical planning a natural testbed for \emph{generalization} across \emph{states}, \emph{goals}, and \emph{problem size} (formal definitions in \cref{sec:background}).

Many challenging learning tasks  in the setting of  classical planning  involve generalization, including
learning a domain from traces drawn from  hidden domain instances  \cite{arora2018review,xi2024neuro,sift}, and closer to the aims
of this work, learning general policies and heuristics  \cite{sylvie:asnet,erez:drl,mausam,sid:generalization,dillon:h}.
A general policy is a policy that can be used to solve
arbitrary instances of the  domain without search, and  a general heuristic is an estimator of the cost to the goal
which can be effectively  used to search for plans in any domain instance. In recent years, DRL approaches
have  been used to learn general policies and heuristics over given domains \cite{erez:drl,simon:kr2023,stahlberg-geffner-aaai2026}.
Yet,  real-time search as used in these  algorithms, moving iteratively  from one state to a successor state \cite{lrta,sven:rts},
is not an  effective way to search for plans.  Planners  use best-first algorithms
like $\wastar{}$ or Greedy BFS \cite{richter-westphal-jair2010,geffner:book,ghallab:book}, and recent approaches have shown indeed the
performance gains that can be  obtained by combining Bellman updates with best-first search,
which is feasible  when the model is known \cite{agostinelli-et-al-nmi2019,orseau-lelis-aaai2021}.

Early algorithms that combine full Bellman updates with real-time search include Learning Real-Time $\astar{}$ ($\lrtastar{}$), for deterministic MDPs,
and Real-Time Dynamic Programming (RTDP), for stochastic MDPs \cite{lrta,rtdp}. RL algorithms, which can be regarded as model-free variants, retain
the real-time search, which is  crucial when interacting with a world or simulator,
but   unnecessary  when the model is known \cite{sutton:book}. In this case, learning can be improved by considering
forms of best-first search, in particular, when the reward (the goal) is sparse. In this work, we push this
idea further by learning to search for plans over the  whole class of instances defined by a given planning domain.
\Omit{
The goal is to achieve a self-improving, generalized learning cycle where, starting with a non-informative heuristic, the search for the goal in one instance  results in a  better heuristic that  aids  the search for the goal in
larger instances, and so on, iteratively.
}
The goal is a self-improving, generalizing learning cycle: starting from a non-informative heuristic, a search for the goal on one instance yields a better heuristic, which then guides search on progressively larger instances.

\begin{figure}[t]
    \centering
    \includegraphics[trim={.9cm 0 .5cm 0}, clip, width=\columnwidth]{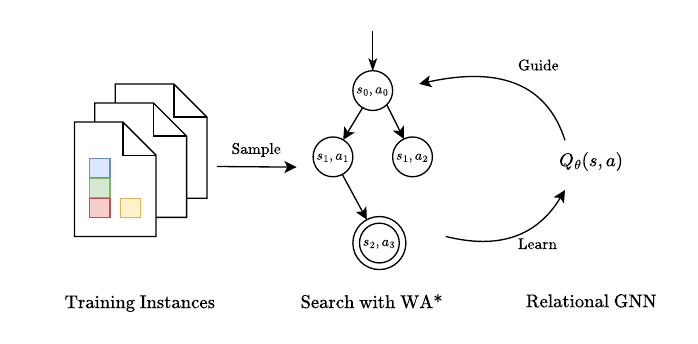} 
    \caption{\textbf{Generalized search for planning.} During training, classical planning instances are solved with a search guided by learned $Q(s,a)$ values, which are then updated to improve performance on other instances. Unlike standard RL, (i) exploration uses best-first search ($\wastar{}$) rather than stepwise real-time search, and (ii) instances vary in both initial state/goal and number of objects.
}
    \label{fig:training_example}
\end{figure}

In the paper, we formulate this form of self-improving generalized search (see \cref{fig:training_example}) that can be applied to families of path-finding problems, 
uniformly modeled as instances of a common planning domain. For this, general $Q$-value functions represented with relational GNNs are learned
by solving training instances with search, using the states deemed relevant for updating the
$Q$-functions. The self-reinforcing
cycle between search and learning appears in a number of research threads in search and RL, but these forms of learning have been tied
to one specific  state space  without generalizing systematically to others. The exceptions are recent works in classical planning for learning
general policies and heuristics, yet the former rely on  standard RL algorithms and hence real-time search \cite{erez:drl,simon:kr2023}
while the latter rely on supervised learning, and do not benefit from the self-improving learning cycle of better heuristics
to guide more efficient searches \cite{dillon-ecai2025,horcik-et-al-aaai25}.

There are three distinct forms of \emph{generalization} in RL and planning: generalization to other \emph{states}, which is required for RL to scale to large state spaces~\cite{silver-et-al-nature2016,silver-et-al-nature2017}; generalization to other \emph{states and goals}, as in goal-conditioned RL~\cite{goal-conditioned-rl,andrychowicz-et-al-nips2017}; and generalization to \emph{states, goals, and problem size}, as pursued in this work.

The paper is organized as follows: we present next the related work, background, and the generalized search for planning framework, followed by our experiments on different benchmarks and the conclusions we draw\footnote{Code and data are available in the project repository: \href{https://github.com/maichmueller/generalized-search-for-planning}{github.com/maichmueller/generalized-search-for-planning}.}.

\section{Related Work}

\paragraph{Planning within RL.}
The Alpha family of RL algorithms \citep{silver-et-al-nature2016, silver-et-al-nature2017, silver-et-al-science2018, schrittwieser-et-al-nature2020} demonstrated the benefits of integrating search into the RL loop. Specifically, Monte Carlo Tree Search (MCTS), guided by value functions and policies learned via actor-critic methods, selects target states for learning, leading to more effective lookaheads and improved learning targets. While this approach performs well in challenging games, MCTS has limitations in our setting: it is ill-suited for single-goal pathfinding, and struggles in non-adversarial puzzle domains where progress depends on sparse or delayed rewards \citep{orseau-lelis-aaai2021}. AlphaZero-based general policy learning is studied in the experiments section as well, and results underline these drawbacks.  

\paragraph{Learning to Search.} Learning Real-Time $\astar{}$ ($\lrtastar{}$) and Real-time Dynamic Programming (RTDP) are aimed at solving deterministic
and stochastic goal-reaching MDPs with a known initial state, combining a greedy real-time search with Bellman updates \cite{lrta,rtdp}.
If the initial heuristics (cost estimators) are admissible, \ie, do not over-estimate true (expected) costs, and there are no dead-end states
from which the goal is not reachable, both $\lrtastar{}$ and RTDP converge to an optimal policy relative to the initial state. More recently, and more closely connected to our work,
different best-first search algorithms have been used inside a learning  algorithm to solve hard  problems like the
$5 \times 5$ sliding puzzle and the Rubik's cube \cite{bootstrap,agostinelli-et-al-nmi2019,orseau-lelis-aaai2021,hadar-et-al-aaai2026}. The results are impressive but their approaches do not address size generalization.

\paragraph{Learning General Policies.}
General policies refer to policies that can solve any (solvable) instance of a given planning domain without search \cite{sylvie:asnet,erez:drl,simon:kr2023}.  Purely symbolic approaches have been developed that result in policies that can be understood and proved correct, but these methods do not scale well \cite{general1,frances-et-al-aaai2021}. 
The most recent works in this thread extend the real-time
search in the DRL loop with an adaptation of hindsight experience replay \cite{andrychowicz-et-al-nips2017}
to the planning setting called \emph{lifted} and \emph{propositional HER}  \cite{stahlberg-geffner-aaai2026},
where the unachieved goals of a trace are replaced by a suitable relational description of substitute goals actually achieved. 
However, this form of HER is most effective when goals can be split into subgoals that can be relabeled; this assumption does not hold uniformly across planning domains and puzzles.
In this paper, we build on this work, but replace the real-time search and HER with a best-first search, which is more effective in hard problems. 

\paragraph{Learning General Heuristics.}
Since there are no perfect general policies for intractable domains like Sokoban, a different research thread focuses on learning general heuristic estimators to guide the search for plans in instances of a given domain. In these works, the heuristics $h(s)$ are learned in a supervised way from optimal values $h^{*}(s)$ precomputed by optimal planners \cite{stahlberg-et-al-icaps2022,horcik-et-al-aaai25,bai-et-al-icaps2025}.
Most approaches use the learned heuristic only to guide search, whereas \citet{bai-et-al-icaps2025} additionally exploits the learned GNN representation for symmetry reduction in best-first search by pruning states through hashed GNN embeddings and actions through approximate graph automorphisms.
To make learned heuristics more cost-effective, some approaches avoid GNNs altogether and instead use efficient support vector machines over relational Weisfeiler--Leman features, which are closely related to the features computable by GNNs \cite{dillon:h}.
Because these methods rely on supervised learning, however, they do not benefit from the self-improving learning cycle of RL approaches.

\paragraph{Search and Exploration in RL.}
The real-time search used in RL algorithms is often extended with bonuses that reward novel states in the search
\cite{explore1,explore2,explore3,explore4}. Indeed, in classical planning, a precise form of novelty, related to a formal notion of problem width, is part of the
state-the-art planning algorithms as well \cite{nir:ecai2012,nir:2017,jendrik:width}. 
Yet all state-of-the-art search algorithms in classical planning
are based on best-first search, not on real-time search where current state is replaced in each step by a successor state \cite{lrta,sven:rts}.

\section{Background}
\label{sec:background}
\paragraph{Classical Planning.}
Classical planning problems are deterministic goal-reaching MDPs with extremely sparse rewards and large state spaces. They are expressed in a \emph{language} (PDDL) that separates the description of the \emph{domain}
$\mathcal{D}$ from the concrete problem \emph{instances}  $\mathcal{I}$ of the domain.
The domain specifies relation types (\emph{predicates}) $\mathcal{P}$, where each predicate
$p\in\mathcal{P}$ has arity $\mathrm{ar}(p)$, as well as \emph{action schemas} $\mathcal{A}$
with lifted  preconditions and effects defined over these relations \cite{geffner:book,ghallab:book,pddl:book}. 
Instantiating a predicate with objects yields \emph{ground atoms}
$p(o_1,\dots,o_{\mathrm{ar}(p)})$, and instantiating an action schema with objects yields a
\emph{ground action} $a(\bar o)$.
An instance $\mathcal{I}$ provides the object set $\mathcal{O}$, the initial state $s_0$, and a goal
specification $g$. The  states  $s$ are represented by sets of  ground atoms; namely, those which
are true in the state. Preconditions and goals are conjunctions of literals, and a goal state
is a state that includes all the goal atoms. A plan is a sequence of applicable actions that maps $s_0$ to a  goal  state.

For example, in  Blocksworld, a 3-block environment is an instance
with objects $\mathcal{O}=\{b_1,b_2,b_3\}$ where the states are described by
means of 4 domain predicates: 
$\pred{on}(\cdot,\cdot)$, $\pred{ontable}(\cdot)$, $\pred{clear}(\cdot)$, and $\pred{holding}(\cdot)$.
The  initial state with  a single tower with $b_3$ at the top, $b_2$ in the middle, and $b_1$ on the table would be
$s_0 = \{\pred{on}(b_3,b_2), \pred{on}(b_2,b_1),\pred{clear}(b_3), \pred{ontable}(b_1)\}$, and the goal
can be given by a single ground atom like $\pred{on}(b_1,b_2)$ or by a conjunction of many such atoms. 
Two of the four action schemas in the domain are 
$\pred{stack}(x,y)$, and $\pred{unstack}(x,y)$, that ground in actions like 
$\pred{stack}(b_1,b_3)$ and $\pred{unstack}(b_3,b_2)$. Only the latter action is applicable in $s_0$.

\Omit{
  An action schema such as $\pred{stack}(x,y)$ has two arguments and is specified by the
precondition  $\{\pred{holding}(x),\pred{clear}(y)\}$ and effects that add $\{\pred{on}(x,y), \pred{clear}(x)\}$ to $s$ and delete $\{\pred{clear}(y), \pred{holding}(x)\}$ from s.
Grounding this schema with objects yields ground actions such as $\pred{stack}(b_3,b_2)$,
which is inapplicable in $s_0$. In turn, $\pred{unstack}(b_3,b_2)$ is applicable in $s_0$ and would satisfy a goal description $g = \{\pred{clear}(b_2)\}$, rendering $\sigma = (\pred{unstack}(b_3,b_2))$ a plan for $g$. \michi{if not wanted here, appendix?}
}

\paragraph{Generalized Planning and Search.}
In generalized planning, we seek a general policy that can solve any domain
instance. These instances vary in the initial state, goals, and number of objects, but the ground actions are instances
of the same action schemas, and the states are all described in terms of the same set of predicates. This is what enables
generalization while defining very precisely the scope of the generalization sought. In domains where learning with
a nearly perfect compact policy is hard or impossible, such as Sokoban \cite{sokoban:np-hard}, search aims at learning heuristics that speed up the search in any instance of the given domain. The approach developed in this paper
serves these two purposes: it can yield nearly-perfect general policies that require no search or informed heuristic estimators.

\paragraph{Search vs. Real-Time search.} In RL and in a number of algorithms like $\lrtastar{}$ and RTDP, one searches for the
goal using real-time search, also called agent-based search \cite{lrta,sven:rts}. In this type of search, there is a
current state $s$  in each iteration such that, in the next iteration, the current state $s'$ is reachable from $s$
by performing one of the applicable actions in $s$. If the \emph{dynamic model of the problem is known}, however, a common
preferred alternative is to search  \emph{best-first} as in $\astar{}$, $\wastar{}$, and GBFS \cite{russell:book}. Best-first search algorithms
explore the space more systematically,  are not affected by dead-ends, and are complete. In each iteration, they all pick the node $n$ from the search boundary that has the minimum evaluation function $f(n)$.
In $\astar{}$, $f(n)=g(n)+h(n)$ where $g(n)$ is the accumulated cost to reach the node $n$ from the root node, in $\wastar{}$,
it is $f(n)=g(n)+wh(n)$ with $w > 1$, giving thus more importance to the estimate of the cost-to-go than to the cost accumulated,
while in greedy best-first search (GBFS)--not to be confused with greedy search (real-time)--the evaluation function is $f(n)=h(n)$.

\section{Generalized Search for Planning}
\label{sec:generalized_search_planning}
In this work, we address exploration in reinforcement learning by relying on \emph{best-first} search, as in classical planning, rather than \emph{real-time} search, as is typical in RL. $\model{}$ (\textbf{G}eneralized \textbf{S}earch for \textbf{P}lanning) is an iterative search-and-learn scheme in which training instances are solved with weighted $\astar{}$ (W$\astar{}$) guided by learned $Q$-values, and the resulting search data is used to improve those $Q$-values. To support generalization across states, goals, and problem sizes, we represent $Q_\theta$ with a relational graph neural network (see \cref{sec:rgnn}).

$\model{}$ maintains a parametric action-value heuristic $Q_\theta(s,a)$ and repeatedly runs W$\astar{}$ on sampled instances to generate experience. Each search episode expands promising state--action pairs, stores encountered transitions in a replay buffer, and (when available) attaches search-derived lower bounds on return. Q-learning updates $Q_\theta$ from this buffer, and the improved $Q_\theta$ in turn guides subsequent search episodes more effectively (cf.\ \cref{fig:training_example}).

\paragraph{Search Episode.}
For a sampled instance $\mathcal{E}$ with initial state $s_0(\mathcal{E})$, we run a W$\astar{}$ search over \emph{state--action} nodes $(s,a)$. The algorithm maintains (i) a search tree rooted at $s_0$, (ii) a priority queue $\mathcal{F}$ (the frontier) containing candidate pairs, and (iii) a replay buffer $\mathcal{D}$ that stores tuples $(s,a,\underline{R})$ whenever a search-derived lower bound $\underline{R}$ on return is available.

We consider unit step rewards $r=-1$ without discounting, so maximizing return corresponds to finding shorter plans. Let $g(s)$ denote the accumulated return of $s$ along the current tree path from $s_0$ to $s$, its negative depth in the tree in our setting. We score each frontier pair by
\begin{equation}
f(s,a) \;=\; g(s) + w\,Q_\theta(s,a), \nonumber
\end{equation}
where $w\in\mathbb{R}$ is a weighting constant. At each expansion, we pop the pair $(s,a)\in\mathcal{F}$ with the highest score and generate the successor state $s' = a(s)$. We distinguish three types of transitions: dead-end, goal, and non-terminal. If the successor $s'$ is a non-terminal state, we insert each \emph{previously unseen} successor pair $(s',a')$ into the tree and push it onto the frontier with score $f(s',a')$. If $s'$ is a dead-end state (no applicable actions), we assign the parent pair $(s,a)$ a fixed penalty return $R_{\bot}$ and store $(s,a,R_{\bot})$ in $\mathcal{D}$.

If $s'$ is a goal state, we backtrack from the goal transition $(s_T,a_T)$ to the root along the discovered solution path and assign each encountered pair $(s_t,a_t)$ the actual return-to-go. This provides a lower bound $\underline{R}(s_t,a_t)$ on the optimal return from that pair, since a higher-return (shorter) solution may exist. We store these pairs together with their lower bounds in $\mathcal{D}$. The search terminates when it reaches a goal state or exhausts its expansion budget. Pseudocode is given in Algorithm \ref{alg:gsp} in the appendix.

\paragraph{Q-learning with Search-Derived Lower Bounds.}
From the replay buffer, we periodically sample batches $(s,a,\underline{R})\sim\mathcal{D}$ and regress $Q_\theta$ toward one-step Bellman targets. Let the bootstrap target be
\begin{equation}
\hat y(s,a) \;=\; -1 + \max_{a' \in \mathcal{A}(s')} Q_\theta(s',a'), \nonumber
\label{eq:bounded-bellman}
\end{equation}
where $s'$ is the successor reached from $(s,a)$. We then set the learning target $y$ by pair type. For non-terminal pairs, we use $y=\hat y(s,a)$ (standard Q-learning), but the search provides two additional supervision signals: dead-end pairs are regressed towards the fixed penalty $y=R_{\bot}$, while goal-path pairs can bound the targets as
\[
y \;=\; \max\{\underline{R},\hat y(s,a)\}.
\]
Since $\underline{R}$ comes from a concrete solution found by search, it lower-bounds the optimal return for $(s,a)$; taking the maximum prevents bootstrap targets from dropping below what search has already achieved and empirically stabilizes learning. Finally, the mean-squared error $\lVert Q_\theta(s,a)-y\rVert^2$ is minimized via stochastic gradient descent.

\paragraph{Instance Selection Strategy.}
Selecting which training instance to solve next is dynamic and consequential: uniform sampling wastes compute on instances that are already solved reliably or are currently out of reach. We therefore maintain three instance pools and sample from them with exponentially increasing weights: \emph{unsolved}, \emph{solved}, and \emph{satisficed}. If a search successfully finds a plan, the instance is placed in \emph{satisficed}. If additionally the expanded-nodes count equals the found plan length, we consider the instance \emph{solved}. Intuitively, instances with suboptimal solutions provide the most informative updates, whereas instances with no solution yet or with near-best solutions contribute less beyond reinforcing existing behavior. Interestingly, if a problem is placed in \emph{solved}, it does not mean it necessarily found the shortest path within the problem, but rather that the heuristic is so confident in guiding the search that no other nodes need to be expanded.

The GSP learning loop dynamics are visualized in Figure \ref{fig:learning_dynamics}, which illustrates the number of training instances the $\wastar{}$ search solved when guided by learned
Q-values, along with the number of nodes expanded in these searches. In the three domains (\textsc{Blocksworld, Satellite, Transport}), it is clear that as learning progresses, more training problems are solved increasingly efficiently within the given search budget. In the fourth domain shown, \textsc{Floortile}, the learning loop in $\model{}$ is not successful in solving all training instances and remains, on average, at high expansion numbers.

\section{Representing $Q_\theta$ with Relational GNNs}
\label{sec:rgnn}
Following earlier works on general policy learning in classical planning \cite{stahlberg-et-al-icaps2022, stahlberg-et-al-ipcl2023, stahlberg-et-al-aaai2025,
aichmueller-geffner-ijcai2025, dillon:neurips24, horcik-et-al-aaai25},
and in particular \cite{stahlberg-geffner-aaai2026}, the Q-function is represented as a relational GNN.
Planning states are indeed relational structures: a state $s$ is a set
of true ground atoms\footnote{We follow the closed-world assumption, \ie, atoms that are not mentioned in a state $s$ are false.} $p(\bar o)$ over a finite object set $\mathcal{O}$, and the goal is
a conjunction $g$ of literals. We write $\bar o=(o_1,\dots,o_{\mathrm{ar}(p)})$ for an ordered tuple of objects matching the arity of predicate $p$. In goal-conditioned RL, both the current state and the goal are required to be encoded. To this end, we form a relation set
\[
\mathcal{R}_{s,g} \coloneqq \{\, p(\bar o) \mid p(\bar o)\in s \cup g \,\}.
\]
Since our heuristic is action-value based, we make action choices explicit during message
passing by augmenting the relational input with auxiliary action-atoms. Concretely, for
each applicable grounded action $a=A(\bar o)\in\mathcal{A}(s)$ we introduce a dedicated
action object $o_a$ (one per applicable action in $s$) and an atom $A(o_a,\bar o)$, and
define
\[
\mathcal{R}_{A} \coloneqq \{\, A(o_a,\bar o) \mid a=A(\bar o)\in \mathcal{A}(s) \,\}.
\]
The final input is the set $\mathcal{R} = \mathcal{R}_{s,g} \cup \mathcal{R}_{A}$
of relational facts over the extended object universe
$\widetilde{\mathcal{O}} = \mathcal{O}\cup\{o_a \colon a \in \mathcal{A}(s)\}$.

We parameterize $Q_\theta(s,a)$ with a relational message-passing network that maintains
a feature embedding $X_i(o)\in\mathbb{R}^d$ for each object $o$ at layer $i$, initialized
to $X_0(o)=\mathbf{0}$. Given $\mathcal{R}$, embeddings are updated for $L$ layers by
exchanging messages along atoms. For each atom
$q = p(o_1,\dots,o_{\mathrm{ar}(p)})\in\mathcal{R}$, a predicate-specific function produces
\emph{position-wise} messages
\begin{equation}
\big(m_{o_1}^q,\dots,m_{o_{\mathrm{ar}(p)}}^q\big)
\;=\;
\mathrm{Comb}_{p}\!\big(X_i(o_1),\dots,X_i(o_{\mathrm{ar}(p)})\big), \nonumber
\label{eq:rgnn:msg}
\end{equation}
so that the message sent to an object depends on its argument role in the relation.
An object $o$ then aggregates all incoming messages across atoms that contain it,
\begin{equation}
m_o \;=\; \mathrm{Agg}\big(\{\, m_{o}^q \;:\; q\in\mathcal{R},\; o\in q \,\}\big), \nonumber
\label{eq:rgnn:agg}
\end{equation}
where $\mathrm{Agg}$ is permutation-invariant. Finally, the embedding is updated with a
shared update function and a residual connection,
\begin{equation}
X_{i+1}(o) \;=\; X_i(o) \;+\; \mathrm{Comb}_{U}\big(X_i(o), m_o\big), \nonumber
\label{eq:rgnn:update}
\end{equation}
for $i=0,\dots,L-1$.

After $L$ layers, we obtain final embeddings $\{X_L(o)\}$ that are equivariant to object
permutations. We compute action-values with a single shared readout that combines the
embedding of the action object with a pooled summary of the state and action objects. Let
\[
\bar X(s,g) \;=\; \mathrm{Pool}\big(\{\, X_L(o) \mid o\in \widetilde{\mathcal{O}}  \,\}\big)
\]
be a permutation-invariant pooling of all embeddings. For an action
$a\in\mathcal{A}(s)$ with associated action object $o_a$, we define
\begin{equation}
Q_\theta(s,a)
\;=\;
\mathrm{MLP}_{Q}\!\Big(\big[\, X_L(o_a) \;\Vert\; \bar X(s,g) \,\big]\Big), \nonumber
\label{eq:q_readout}
\end{equation}
where $[\cdot \Vert\cdot]$ denotes concatenation.
Importantly, $\mathrm{MLP}_{Q}$ is shared across all action schemas; action types and
argument structure are expressed through the relational message passing and the resulting
embedding $X_L(o_a)$. This way, the model learns a common scoring principle for grounded actions instead of separate schema-specific predictors.

We use smoothmax aggregation and parameterize the update function $\mathrm{Comb}_{U}$ with
a single shared multi-layer perceptron (MLP) mapping $\mathbb{R}^{2d}$ to $\mathbb{R}^{d}$.
In contrast, we implement a separate predicate-specific MLP for each $\mathrm{Comb}_{p}$,
\ie, one MLP per predicate symbol $p\in\mathcal{P}$, mapping
$\mathbb{R}^{d \cdot \mathrm{ar}(p)}$ to $\mathbb{R}^{d \cdot \mathrm{ar}(p)}$ and
interpreting the output as $\mathrm{ar}(p)$ many position-wise messages in $\mathbb{R}^d$.
\section{Experiments}
\label{sec:experiments}

Our experiments evaluate $\model{}$ along two axes: \emph{generalization} to unseen planning
instances (varying in size, initial state, and goal) and \emph{exploration efficiency}
during training. During training, $\model{}$ generates experience via heuristic-guided
weighted $\astar{}$ ($w=2$). At test time, the learned $Q_\theta$ can be used to define a greedy policy or to guide a best-first search. We therefore report results for both greedy execution (\textsc{GSP}$_\pi$) and $\wastar{}(w=2)$ guided by $Q_\theta$ ($\model{}_{\wastar{}}$).

We consider three types of domains: planning domains, puzzles, and the PushWorld domain, all described below.
Beyond final coverage and plan-quality metrics, we analyze training dynamics by tracking the number of node expansions and instance solve rate over training time, indicating whether learning yields increasingly focused search. Figure \ref{fig:learning_dynamics} shows four exemplary domains, with the remaining domain plots found in the appendix.

\begin{figure}[ht]
    \centering
    \includegraphics[width=\columnwidth]{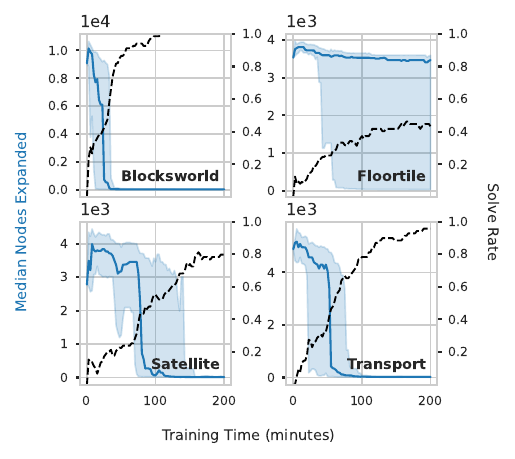} 
    \caption{Training progress across selected IPC-learning domains. The blue line (left axis) shows the median number of expanded nodes, with shaded regions representing the 25th and 75th percentiles across all considered training instances. The black line (right axis) shows the solve rate, indicating the percentage of training instances where a goal has been found during training. Each training run lasted 12 hours (720 minutes) of which we show the first 180 minutes here.}
    \label{fig:learning_dynamics}
\end{figure}

\paragraph{Training Setup.}
\label{sec:exp-training}

The same hyperparameters were used in all experiments. In particular, we use an embedding dimension $d=32$ and smooth-maximum aggregation.
The learning rate is set to $10^{-4}$ for the R-GNN parameters and $10^{-3}$ for the readout network. Training is parallelized with one learner process and five search workers that generate experience concurrently.
Each search episode uses a 60\,s expansion budget and worker-side batching with batch size 256. We use a FIFO replay buffer with a capacity of 40 batches. We use a target network for Q-learning and update it every 10 passes through the replay buffer \cite{mnih-et-al-nature2015}. A detailed overview of how we do final model selection is in the appendix.

\subsection{Planning Domains}
\label{sec:exp-planning-domains}

We use the planning benchmarks from the 2023 International Planning Competition (IPC) learning track. The benchmark comprises 10 domains, each providing 100 training instances
(70 used for training and 30 for validation) and 90 test instances. 
The learning-track instances were designed to stress domain-independent planners by
inducing large increases in object counts and applicable actions. For example, in
\textsc{Childsnack}, the hardest test instance admits 46{,}703{,}541 applicable actions in
the initial state, making unguided search infeasible and placing strong pressure on
heuristic quality. An overview of scaling is provided in
\cref{tab:scaling}.

\begin{table}[h]
\centering
\begin{tabular}{l l r}
\toprule
\textbf{Domain} & \textbf{Objects} (train/test) & \textbf{Plan} (train/test) \\
\midrule
Blocks      & \planpairl{$29$}{$488$} & \planpair{$102$}{$1786$} \\
Transport   & \planpairl{$34$}{$453$} & \planpair{$57$}{$1083$} \\
Sokoban     & \planpairl{$11\!\times\!11$}{$99\!\times\!99$ $(b=3/79)$} & \planpair{$22$}{$10546$} \\
Spanner     & \planpairl{$28$}{$833$} & \planpair{$80$}{$831$} \\
Childsnack  & \planpairl{$51$}{$1326$} & \planpair{$33$}{$879$} \\
Satellite   & \planpairl{$27$}{$596$} & \planpair{$41$}{$3428$} \\
Floortile   & \planpairl{$9\!\times\!3$}{$34\!\times\!28$ $(r=2/26)$} & \planpair{$105$}{$3398$} \\
Miconic     & \planpairl{$21$}{$681$} & \planpair{$24$}{$1361$} \\
Ferry       & \planpairl{$25$}{$1461$} & \planpair{$51$}{$3895$} \\
Rovers      & \planpairl{$28$}{$596$} & \planpair{$41$}{$3428$} \\
\bottomrule
\end{tabular}
\caption{Scaling of problem size across domains. \textbf{Objects (train/test)} reports the largest number of objects in any instance of the training and test splits (or domain-specific size parameters for grid domains where applicable: Sokoban uses grid size and boxes $b$, Floortile grid size and robots $r$). \textbf{Plan (train/test)} reports the number of steps in the (suboptimal) example solutions provided by the 2023 IPC learning benchmark. \emph{These are not used as bounds for training}.}
\label{tab:scaling}
\end{table}

\paragraph{Baselines.}
We compare against \textsc{Lifted HER} \cite{stahlberg-geffner-aaai2026}, a method similar to ours that exploits the relational structure using R-GNNs, while learning with real-time search, hindsight relabeling, and DQN \cite{mnih-et-al-nature2015} updates. We further compare with the
domain-independent planner \textsc{LAMA} \cite{richter-westphal-jair2010}, and report
published solve rates for WL features (WL-f) \cite{dillon-ecai2025}, the state encoding of
Horc\'{\i}k et al.~\cite{horcik-et-al-aaai25}, and Distincter \citep{bai-et-al-icaps2025}. We additionally evaluate an AlphaZero ($\alpha_0$) policy-value baseline using the same relational network architecture as $\model{}$. Instead of generating data with $\wastar{}$, it runs Monte Carlo Tree Search from the current state, using a network policy $\pi_\theta( \cdot \mid s)$ as a prior over applicable grounded actions and the value head to bootstrap newly expanded leaves. Root visit counts provide the policy target, and the value head is trained on the undiscounted return of the executed episode suffix. We execute the learned policy as a greedy search at test time. This baseline contrasts best-first search with the local MCTS as learning driver for generalized planning.

$\model{}$ is evaluated in two test-time
modes: greedy execution induced by $Q_\theta$ (\textsc{GSP}$_\pi$) and $\wastar{}$ ($w=2$) guided by
$Q_\theta$ (\textsc{GSP}$_{\wastar{}}$).
A secondary experiment establishes a head-to-head comparison with \textsc{Lifted HER} on the benchmarks used in their evaluation. Details on this can be found in the appendix.

\paragraph{Results.}

\begin{table*}[t]
  \centering
  \setlength{\tabcolsep}{5.85pt}
  \begin{tabular}{l|cc|cc|cc|cc|c|c|c|c}
  \toprule
  Domain
  & \multicolumn{2}{c|}{GSP$_\pi$}
  & \multicolumn{2}{c|}{GSP$_{\wastar{}}$}
  & \multicolumn{2}{c|}{Lifted HER}
  & \multicolumn{2}{c|}{LAMA}
  & WL-f
  & Horc\'{\i}k
  & Distincter
  & $\alpha_0$ \\
  &
  \multicolumn{1}{>{\columncolor{covbg}}c}{Cov.} & Steps
  & \multicolumn{1}{>{\columncolor{covbg}}c}{Cov.} & Steps
  & \multicolumn{1}{>{\columncolor{covbg}}c}{Cov.} & Steps
  & \multicolumn{1}{>{\columncolor{covbg}}c}{Cov.} & Steps
  & \multicolumn{1}{>{\columncolor{covbg}}c|}{Cov.}
  & \multicolumn{1}{>{\columncolor{covbg}}c|}{Cov.} 
  & \multicolumn{1}{>{\columncolor{covbg}}c|}{Cov.} 
  & \multicolumn{1}{>{\columncolor{covbg}}c}{Cov.} \\
  \midrule
  blocksworld
  & \textcolor{covgreen!100!covred}{100\%} & 444
  & \textcolor{covgreen!79!covred}{79\%} & 240
  & \textcolor{covgreen!98!covred}{98\%} & 421
  & \textcolor{covgreen!61!covred}{61\%} & 303
  & \textcolor{covgreen!73!covred}{73\%}
  & \textcolor{covgreen!59!covred}{59\%} 
  & \textcolor{covgreen!98!covred}{98\%} 
  & \textcolor{covgreen!31!covred}{31\%} \\
  childsnack
  & \textcolor{covgreen!41!covred}{41\%} & 31
  & \textcolor{covgreen!29!covred}{29\%} & 22
  & \textcolor{covgreen!40!covred}{40\%} & 35
  & \textcolor{covgreen!40!covred}{40\%} & 46
  & \textcolor{covgreen!52!covred}{52\%}
  & - 
  & \textcolor{covgreen!71!covred}{71\%}
  & \textcolor{covgreen!0!covred}{0\%} \\
  ferry
  & \textcolor{covgreen!87!covred}{87\%} & 422
  & \textcolor{covgreen!77!covred}{77\%} & 232
  & \textcolor{covgreen!100!covred}{100\%} & 736
  & \textcolor{covgreen!78!covred}{78\%} & 285
  & \textcolor{covgreen!73!covred}{73\%}
  & \textcolor{covgreen!66!covred}{66\%} 
  & \textcolor{covgreen!92!covred}{92\%}
  & \textcolor{covgreen!19!covred}{19\%} \\
  floortile
  & \textcolor{covgreen!20!covred}{20\%} & 54
  & \textcolor{covgreen!28!covred}{28\%} & 57
  & \textcolor{covgreen!0!covred}{0\%} & - 
  & \textcolor{covgreen!13!covred}{13\%} & 47
  & \textcolor{covgreen!3!covred}{3\%}
  & \textcolor{covgreen!32!covred}{32\%} 
  & \textcolor{covgreen!2!covred}{2\%} 
  & \textcolor{covgreen!0!covred}{0\%} \\
  miconic
  & \textcolor{covgreen!100!covred}{100\%} & 490
  & \textcolor{covgreen!98!covred}{98\%} & 268
  & \textcolor{covgreen!90!covred}{90\%} & 562
  & \textcolor{covgreen!100!covred}{100\%} & 301
  & \textcolor{covgreen!98!covred}{98\%}
  & \textcolor{covgreen!68!covred}{68\%} 
  & \textcolor{covgreen!100!covred}{100\%} 
  & \textcolor{covgreen!28!covred}{28\%} \\
  rovers
  & \textcolor{covgreen!24!covred}{24\%} & 380
  & \textcolor{covgreen!11!covred}{11\%} & 17
  & \textcolor{covgreen!32!covred}{32\%} & 267
  & \textcolor{covgreen!79!covred}{79\%} & 277
  & \textcolor{covgreen!50!covred}{50\%}
  & \textcolor{covgreen!33!covred}{33\%} 
  & \textcolor{covgreen!47!covred}{47\%} 
  & \textcolor{covgreen!0!covred}{0\%} \\
  satellite
  & \textcolor{covgreen!61!covred}{61\%} & 134
  & \textcolor{covgreen!33!covred}{33\%} & 18
  & \textcolor{covgreen!56!covred}{56\%} & 394
  & \textcolor{covgreen!100!covred}{100\%} & 163
  & \textcolor{covgreen!57!covred}{57\%}
  & \textcolor{covgreen!40!covred}{40\%} 
  & \textcolor{covgreen!53!covred}{53\%} 
  & \textcolor{covgreen!0!covred}{0\%} \\
  sokoban
  & \textcolor{covgreen!14!covred}{14\%} & 18
  & \textcolor{covgreen!32!covred}{32\%} & 30
  & \textcolor{covgreen!8!covred}{8\%} & 11
  & \textcolor{covgreen!44!covred}{44\%} & 143
  & \textcolor{covgreen!37!covred}{37\%}
  & \textcolor{covgreen!30!covred}{30\%} 
  & \textcolor{covgreen!36!covred}{36\%} 
  & \textcolor{covgreen!0!covred}{0\%} \\
  spanner
  & \textcolor{covgreen!100!covred}{100\%} & 216
  & \textcolor{covgreen!27!covred}{27\%} & 14
  & \textcolor{covgreen!97!covred}{97\%} & 160
  & \textcolor{covgreen!33!covred}{33\%} & 14
  & \textcolor{covgreen!71!covred}{71\%}
  & \textcolor{covgreen!59!covred}{59\%} 
  & \textcolor{covgreen!100!covred}{100\%}
  & \textcolor{covgreen!73!covred}{73\%} \\
  transport
  & \textcolor{covgreen!73!covred}{73\%} & 448
  & \textcolor{covgreen!57!covred}{57\%} & 52
  & \textcolor{covgreen!96!covred}{96\%} & 228
  & \textcolor{covgreen!77!covred}{77\%} & 86
  & \textcolor{covgreen!56!covred}{56\%}
  & \textcolor{covgreen!39!covred}{39\%} 
  & \textcolor{covgreen!56!covred}{56\%} 
  & \textcolor{covgreen!0!covred}{0\%} \\
  \bottomrule
  \end{tabular}
    \caption{Results on 2023 IPC benchmark. We report coverage ('Cov.' fraction of 90 test instances solved) and average plan length (Steps) over solved instances for $\model{}$ as a greedy policy ($\model{}_\pi$) and as a $\wastar{}(w=2)$ heuristic ($\model{}_{\wastar{}}$), alongside \textsc{Lifted HER} and \textsc{LAMA}. Published coverage for WL features (WL-f) \cite{dillon-ecai2025} and Horc\'{\i}k et al. \cite{horcik-et-al-aaai25} is shown for reference. The $\model{}$ budget for each instance is limited to 10,000 expansions or one hour, whichever occurs first.}
    \label{tab:ipc-results}
\end{table*}

\cref{tab:ipc-results} summarizes results on IPC 2023 learning-track domains.
Greedy execution performs strongly on several domains: \textsc{GSP}$_\pi$ reaches
$100\%$ coverage on \textsc{Blocksworld}, \textsc{Miconic}, and \textsc{Spanner}, surpassing all baselines, and
achieves competitive coverage on \textsc{Ferry} ($87\%$) and \textsc{Transport} ($73\%$), where only \textsc{Lifted HER} can outperform in both. Against WL-f and Horc\'{\i}k et al., \textsc{GSP}$_\pi$ is particularly
strong on \textsc{Blocksworld} and \textsc{Miconic}, while the alternative baselines are
stronger on \textsc{Floortile} and sometimes on \textsc{Childsnack}.

Using $Q_\theta$ as a $\wastar{}$ heuristic yields mixed outcomes.
In some domains, $\wastar{}$ improves coverage over greedy execution, most notably in the puzzle domains \textsc{Sokoban} ($14\%\!\rightarrow\!32\%$) and \textsc{Floortile} ($20\%\!\rightarrow\!28\%$). In contrast, $\wastar{}$ reduces coverage on several domains where greedy execution is already strong (\eg, \textsc{Blocksworld},
\textsc{Spanner}, and \textsc{Transport}), indicating that the benchmark's large branching factors can cause failures, even with effective heuristics.

The two domains, \textsc{Rovers} and
\textsc{Satellite}, are solved reliably only by \textsc{LAMA} with 79\% and 100\% coverage,
respectively. This is an expected result due to known limitations of 1-WL expressivity on these domains, which affects all learning-based methods, but not a domain-independent planner like \textsc{LAMA} \cite{drexler-et-al-kr2024, horcik-et-al-icaps-2024}.
\textsc{Childsnack} is challenging for all methods, which we attribute primarily to the extreme branching factors discussed above. Distincter is the only method that performs well in this domain, suggesting that symmetry pruning partially counteracts the large branching factor.

In contrast, AlphaZero-style learning did not achieve sufficient generalization, with consistently weak results across domains except in Spanner. This lack of generalization was already visible during training, where runs often failed to generalize to all validation instances. These results support our hypothesis that MCTS is poorly suited as the main search mechanism for general policy learning in this setting. However, we do not exclude that a more targeted study on learning general policies with AlphaZero -- particularly one that provides larger simulation budgets, more efficient implementations, and further algorithmic improvements from recent work on AlphaZero -- may obtain stronger performance.

The difference in performance between $\model{}_{\pi}$ and $\model{}_{\wastar{}}$ on the IPC 2023 benchmark is attributable to two reasons. Firstly, huge branching factors in test instances degrade search, as the benchmark's design challenges search algorithms specifically. Secondly, the learned heuristic is substantially stronger as a local ranking mechanism than as a global value function on out-of-distribution data. On training problems, the learned heuristic eventually scores well enough globally that the number of expanded nodes is close to or equal to the solution length, indicating that the search is perfectly guided. However, this property deteriorates more quickly than the relative ranking of applicable actions at a state when generalizing to larger or structurally different test instances. As a result, greedy search can remain effective because it relies only on choosing the best local action, whereas $\wastar{}$ must order a frontier using poorly calibrated global scores. This behavior is due to encoding the state together with all applicable actions into a joint relational graph, allowing the GNN to score the actions in direct context to one another, which directly benefits action selection. However, this design does \emph{not} encourage well-calibrated scores across states.

\subsection{Puzzles}
\label{sec:exp-puzzles}

\begin{table*}[t]
  \centering
  \begin{small}
  \begin{sc}
  \begin{tabular}{l c c c c}
    \toprule
        Domain $\longrightarrow$ & \multicolumn{4}{c}{\textbf{\textcolor{sokcol}{Sokoban} / \textcolor{witcol}{The Witness} / \textcolor{tilecol}{Sliding Tile Puzzle 5x5}}} \\
    \midrule
Model & Solved & Length & Expansions & Time (s) \\
    \midrule
    \midrule
    \textsc{GSP}$_\pi$
      & \triple{681}{667}{0}
      & \triple{33.7}{16.3}{--}
      & \wtriple{34}{16}{--}
      & \triple{3.8}{3.0}{--} \\
    \textsc{GSP}$_{\mathrm{GBFS, b=1}}$
      & \triple{998}{1000}{0}
      & \triple{38.5}{16.2}{--}
      & \wtriple{564}{496}{--}
      & \triple{59.7}{84.0}{--} \\
      \textsc{GSP}$_{\mathrm{GBFS, b=32}}$
      & \triple{1000}{1000}{--}
      & \triple{32.6}{14.7}{--}
      & \wtriple{1028}{722}{--}
      & \triple{103}{72.0}{--} \\
    \textsc{GSP}$_{\wastar{}, w=2, b=1}$
      & \triple{1000}{1000}{0}
      & \triple{36.0}{16.0}{--}
      & \wtriple{207}{548}{--}
      & \triple{22.1}{94.2}{--} \\
    \textsc{GSP}$_{\wastar{}, w=2, b=32}$
      & \triple{1000}{1000}{--}
      & \triple{32.5}{14.7}{--}
      & \wtriple{972}{765}{--}
      & \triple{61.1}{78.7}{--} \\
    \midrule

    \textsc{Lifted HER}$_\pi$ 
      & \triple{309}{-}{0}
      & \triple{31.2}{--}{--}
      & \wtriple{31}{--}{--}
      & \triple{--}{--}{--} \\
    GBFS ($\dagger$)
      & \triple{914}{290}{0}
      & \triple{37.7}{13.3}{--}
      & \wtriple{5040}{10128}{--}
      & \triple{49.2}{44.6}{--} \\
    $\wastar{}, w=2$ ($\dagger$)
      & \triple{1000}{835}{1000}
      & \triple{35.6}{14.2}{130.3}
      & \wtriple{3298}{14305}{1802}
      & \triple{22.8}{55.5}{1.5} \\
    $\mathrm{PHS}^*$ ($\dagger$)
      & \triple{1000}{1000}{1000}
      & \triple{37.6}{14.4}{222.8}
      & \wtriple{1522}{191}{2764}
      & \triple{11.3}{1.7}{3.0} \\
    $\mathrm{LevinTS}$ ($\dagger$)
      & \triple{1000}{1000}{30}
      & \triple{40.1}{14.8}{159.6}
      & \wtriple{2640}{220}{65545}
      & \triple{19.5}{1.6}{56.7} \\
    $\mathrm{PHS}_h$ ($\dagger$)
      & \triple{1000}{1000}{4}
      & \triple{38.9}{14.6}{119.5}
      & \wtriple{1962}{222}{58692}
      & \triple{14.8}{1.8}{55.3} \\
    DeepCubeA ($\square$)
      & \triple{1000}{--}{--}
      & \triple{32.88}{--}{--}
      & \wtriple{1050}{--}{--}
      & \triple{--}{--}{--} \\
    LAMA
      & \triple{1000}{--}{--}
      & \triple{51.60}{--}{--}
      & \wtriple{3150}{--}{--}
      & \triple{--}{--}{--} \\
    \bottomrule
  \end{tabular}
  \end{sc}
  \end{small}
    \caption{Performance comparison on puzzle domains Sokoban ($10\times 10$, 4 boxes) / The Witness ($5\times 5$) / Sliding Tile ($5\times 5$). 'Solved' is the number of solved instances out of 1000 per domain. Length, Expansions, and Time report averages over the solved instances in each domain ('--' if results unavailable). Sokoban models were trained on the same training problems as \citet{orseau-lelis-aaai2021} and DeepCubeA \cite{agostinelli-et-al-nmi2019}. All domains are converted to PDDL with relational encoding. LAMA results are taken from \citet{agostinelli-et-al-nmi2019} where available. The symbol $\dagger$ refers to  \citet{orseau-lelis-aaai2021}, while $\square$ refers to \citet{agostinelli-et-al-nmi2019}. The $\model{}$ budget for each instance is limited to 100,000 expansions or five hours, whichever occurs first. We ran search modes with batch size $b=1$ and $b=32$, respectively, allowing a fairer comparison against baselines that reported results with $b=32$.}
  \label{tab:puzzles-3way}
\end{table*}

We next consider combinatorial puzzles that primarily test \emph{structural} (same-size) generalization: \textsc{24-Puzzle}, \textsc{Sokoban} ($10\times 10$, 4 boxes), and \textsc{The Witness} ($5\times 5$), following \citet{orseau-lelis-aaai2021}. We use the same training and test splits, but evaluate in a relational setting by converting the environments to PDDL. 
In contrast to IPC domains, instance sizes are fixed (or vary only mildly); generalization requires transferring relational reasoning to unseen configurations. Notably, unlike the IPC \textsc{Sokoban} domain, this version allows arbitrary assignment of boxes to goal locations.

\paragraph{Baselines.}
We compare against established puzzle solvers from \citet{orseau-lelis-aaai2021}, including
GBFS, $\wastar{}(w=2)$, and Levin's policy-guided tree-search variants (LevinTS, $\mathrm{PHS}^*$,
$\mathrm{PHS}_h$). We also report results for \textsc{DeepCubeA} on \textsc{Sokoban}
\cite{agostinelli-et-al-nmi2019} and \textsc{LAMA} where available. For $\model{}$, we report the modes $\model{}_\pi$, $\model{}_{\wastar{}}$, and $\model{}_{\mathrm{GBFS}}$.

\paragraph{Results.}
\cref{tab:puzzles-3way} summarizes results across all three puzzle domains.
While two methods from \citet{orseau-lelis-aaai2021}, $\wastar{}$ and PHS$^*$, were able to solve all of the \textsc{24-Puzzle} test instances, $\model{}$ did not find a single goal path during training, and subsequently fails at test-time. Finding a solution to one of the training instances is challenging in this domain, as the training instances are not easy and initial $Q_\theta$ values are not informed. The successful baselines increase their search budgets upon failure during training and, paired with more efficient fixed-size MLPs instead of size-adaptive relational GNNs, are able to eventually find learning signals, while $\model{}$ on a fixed budget cannot. Likewise, \textsc{Lifted HER} is not able to learn successfully.

On \textsc{Sokoban}, \textsc{GSP}$_\pi$ solves 681/1000 instances, more than double the
coverage of \textsc{Lifted HER}$_\pi$ (309). When used for search, $\model{}_{\wastar{}}$ yields perfect performance, solving all instances with 207 expansions on average, compared to 3298 expansions for $\wastar{}(w=2$) in \citet{orseau-lelis-aaai2021} and 1050 expansions for \textsc{DeepCubeA}. Similarly, $\model{}_{\mathrm{GBFS}}$ reduces the required expansions from 5040 (GBFS baseline) to 564, while simultaneously increasing coverage from 914 to 998/1000.

On \textsc{The Witness}, both $\model{}_{\wastar{}}$ and $\model{}_{\mathrm{GBFS}}$ solve all instances, with expansions in the same range as LevinTS and PHS variants. This pattern is consistent with the importance of dead-end avoidance in this domain: $\model{}$ is trained not only from goal-reaching trajectories but also from explicit dead-end states. $\model{}_\pi$ solves 667 instances, which shows that the learned heuristic is able to surpass even the GBFS baseline (290) in this domain, while losing only to their $\wastar{}$ search (835).  Results for \textsc{DeepCubeA} and \textsc{Lifted HER} are not available for \textsc{The Witness} in the cited works, and adapting their training procedures to this setting is non-trivial (reverse walks and goal relabeling depend on unavailable representation structure).

\subsection{PushWorld}
\label{sec:exp-pushworld}

PushWorld is a Sokoban-like benchmark with sequential pushing and additional object types like composite shapes, and non-goal objects \cite{kansky2023pushworld}. The benchmark is organized into Levels~0--5: Level~0 instances are procedurally generated, whereas Levels~1--5 are hand-designed. We use a custom PDDL formulation (released with our code).
We train on Level~0 and evaluate transfer to Level~1, for which it is substantially harder to find generalizing behavior due to highly varying and larger multi-box shapes and the resulting long-horizon spatial reasoning.

\paragraph{Baselines.}
We compare $\model{}_\pi$, $\model{}_{\mathrm{GBFS}}$, and $\model{}_{\wastar{}}$ against the model-free RL baselines reported by \citet{kansky2023pushworld} (DQN, PPO) and against \textsc{LAMA} \cite{richter-westphal-jair2010}.

\paragraph{Results.}
\cref{tab:pushworld-2way} summarizes results. As a greedy policy, $\model{}_\pi$ solves
93/200 Level~0 test instances, substantially outperforming DQN (20/200) and PPO (11/200), while producing short plans on the solved subset. When used for search, $\model{}_{\wastar{}}$ solves all Level~0 instances (200/200) matching the performance of \textsc{LAMA} in coverage, while yielding higher quality plans on average (18 vs. 24 steps). $\model_{\mathrm{GBFS}}$ solves one instance fewer, but also creates worse plans on average (27 steps) than the $\wastar{}$ variant and LAMA.

Transfer to Level~1 remains challenging, but $Q_\theta$ still provides effective search guidance without requiring fine-tuning: 
\textsc{GSP}$_{\wastar{}}$ solves 48/63 evaluable instances, with plan lengths again shorter than those of \textsc{LAMA} (26 vs 36 steps) on those instances solved by both methods. Yet, it requires roughly $30\times$ fewer expansions to do so. Again, $\model{}_{\mathrm{GBFS}}$ performs slightly worse, solving 4 instances fewer and with an overall worse plan quality of 47 steps.

\begin{table}[h]

  \centering
  \setlength{\tabcolsep}{5.5pt}
  \begin{small}
  \begin{sc}
  \begin{tabular}{l c c c c}
    \toprule
    Domain & \multicolumn{4}{c}{\textbf{\textcolor{sokcol}{Level\,0$_{\textsc{All}}$} / \textcolor{witcol}{Level\,1}}} \\
    \midrule
    Model & Solved & Length & Expanded & Time (s) \\
    \midrule
    \textsc{GSP}$_\pi$ 
      & \double{93}{5}
      & \double{13}{24}
      & \double{13}{24}
      & \double{2}{8} \\
    \textsc{GSP}$_{\mathrm{GBFS}}$ 
      & \double{199}{44}
      & \double{27}{47}
      & \double{1.2k}{3.6k}
      & \wdouble{183}{1314} \\
    \textsc{GSP}$_{\wastar{}}$ 
      & \double{200}{48}
      & \double{18}{26}
      & \double{1.8k}{3.8k}
      & \double{273}{576} \\
    \midrule
    PPO 
      & \double{11}{4}
      & \double{--}{--}
      & \double{--}{--}
      & \double{--}{--} \\
    DQN
      & \double{20}{$\approx 1$}
      & \double{--}{--}
      & \double{--}{--}
      & \double{--}{--} \\
    LAMA 
      & \double{200}{60}
      & \double{24}{40}
      & \wdouble{6.9k}{118k}
      & \double{2}{41} \\
    \bottomrule
  \end{tabular}
  \end{sc}
  \end{small}
\caption{PushWorld test results on Level\,0$_{\textsc{All}}$ (200) and Level\,1 (68). Five Level\,1 instances are removed due to requiring a predicate previously unseen during training on level\,0. $\model{}$ is trained on Level\,0 and evaluated on both levels. 'Solved' reports solved instances; 'Length', 'Expanded', and 'Time' are averages over solved instances ('--' if unavailable), all rounded to integers. LAMA is given 90\,GB memory- and 30 minute time budgets per instance. PPO and DQN results are from \citet{kansky2023pushworld}. The $\model{}$ budget for each instance is limited to 100,000 expansions or five hours, whichever occurs first.}
  \label{tab:pushworld-2way}
\end{table}

\subsection{Ablations}
\label{sec:exp-ablations}

We evaluate the contribution of each training component through three ablations of $\model{}$ on the IPC planning domains: removing (i) dead-end supervision, (ii) solution lower-bound targets, and (iii) the priority-based instance sampling strategy. For each ablation, we train five seeds and evaluate the best checkpoint from each seed according to validation performance. We then report mean and standard deviation over the corresponding test results. The full results are shown in \cref{tab:ipc-2023-ablations-gsp} in the appendix.

Overall, each ablated variant performs worse than the complete model on most domains, although the degradation is usually moderate. The trend is more pronounced for $\model{}_\pi$ than for $\model{}_{\wastar{}}$. This suggests that the search-based evaluator can partially compensate for weaker learned guidance, while the hardest domains also create floor effects that make ablation differences less visible. Among the components, dead-end supervision has the smallest impact, which is consistent with only a few IPC domains exhibiting dead-ends. Removing solution lower-bound targets causes the largest degradations, indicating that the target bounds aid in avoiding accidental regressions. The priority-based sampling strategy has a smaller effect on mean performance. More generally, the ablations show that similar results can be obtained without each component, but the complete model produces the most stable performance across domains and random seeds.

\section{Conclusion}

We introduced a simple framework for learning general $Q$-functions to guide the search for plans on arbitrary instances of a given classical planning domain.
The $Q$-functions, represented by relational GNNs, are learned from instances by performing a best-first ($\wastar{}$) search, informed by the $Q$-values themselves.
This creates a self-improving loop in which search provides training targets and the learned heuristic improves subsequent searches.
The resulting $Q$-functions generalize to other domain instances with different states, goals, and numbers of objects, and can be used at test time either as greedy policies or as heuristics for search.

The experiments cover a broad set of domains, ranging from the large, highly scaled instances of the 2023 International Planning Competition to bounded-size combinatorial puzzles and the challenging PushWorld benchmark used to evaluate RL and classical planning algorithms.
The performance of $\model{}$ competes with the state of the art in almost all of these domains, while addressing this range of problems in a uniform manner, using the same architecture and hyperparameters.
The results suggest that best-first search is a useful exploration mechanism for generalized reinforcement learning when the transition model is known, particularly in sparse-reward domains where real-time search struggles to discover informative trajectories.

At the same time, a limitation of the approach is that the learned $Q$-function is often stronger as a local action-ranking mechanism than as a globally calibrated heuristic over the full search frontier.
Improving value generalization across states and combining best-first search with the complementary strengths of real-time search and HER are natural directions for future work.


\section*{Impact Statement}

This paper presents work whose goal is to advance the field of Machine
Learning. There are many potential societal consequences of our work, none of
which we feel must be specifically highlighted here.
\clearpage
\section*{Acknowledgements}
The research has been supported by the Alexander von Humboldt Foundation with funds from the Federal Ministry for Education and Research, Germany.
This project has received funding from the European Research Council (ERC) under the European Union's Horizon 2020 research and innovations programme (Grant agreement No. 885107).
This project was also funded by the German Federal Ministry of Education and Research (BMBF) and the Ministry of Culture and Science of the German State of North Rhine-Westphalia (MKW) under the Excellence Strategy of the Federal Government and the L\"ander.
\bibliography{bibliography,control}

@String{jair = "Journal of Artificial Intelligence Research"}

@String{nature = "Nature"}

@String{symmetry = "Symmetry"}

@String{nmi = "Nature Machine Intelligence"}

@article{agostinelli-et-al-nmi2019,
  author  = {Forest Agostinelli and Stephen McAleer and Alexander Shmakov and Pierre Baldi},
  title   = {Solving the {Rubik}'s cube with deep Reinforcement Learning and search},
  journal = nmi,
  volume  = {1},
  pages   = {356--363},
  year    = {2019},
  doi     = {10.1038/s42256-019-0070-z}
}

@inproceedings{drexler-et-al-kr2024,
  author    = {Dominik Drexler and Simon St{\aa}hlberg and Blai Bonet and Hector Geffner},
  title     = {Symmetries and Expressive Requirements for Learning General Policies},
  booktitle = {Proceedings of the 21st International Conference on Principles of Knowledge Representation and Reasoning (KR 2024)},
  year      = {2024}
}

@inproceedings{frances-et-al-aaai2021,
  author    = {Guillem Franc{\`e}s and Blai Bonet and Hector Geffner},
  title     = {Learning General Planning Policies from Small Examples Without Supervision},
  booktitle = {Proceedings of the 35th {AAAI} Conference on Artificial Intelligence ({AAAI} 2021)},
  year      = {2021},
  pages     = {11801--11808}
}

@article{orseau-lelis-aaai2021, 
    title={Policy-Guided Heuristic Search with Guarantees}, 
    volume={35}, 
    url={https://ojs.aaai.org/index.php/AAAI/article/view/17469},
    DOI={10.1609/aaai.v35i14.17469}, 
    number={14},
    journal={Proceedings of the AAAI Conference on Artificial Intelligence}, 
    author={Orseau, Laurent and Lelis, Levi H. S.}, year={2021},
    month={May}, 
    pages={12382–12390}
}

@article{richter-westphal-jair2010,
  author  = {Silvia Richter and Matthias Westphal},
  title   = {The {LAMA} Planner: Guiding Cost-Based Anytime Planning with Landmarks},
  journal = jair,
  volume  = {39},
  year    = {2010},
  pages   = {127--177}
}

@article{silver-et-al-nature2016,
  author  = {David Silver and Aja Huang and Chris J. Maddison and Arthur Guez and Laurent Sifre and George {van den Driessche} and Julian Schrittwieser and Ioannis Antonoglou and Veda Panneershelvam and Marc Lanctot and Sander Dieleman and Dominik Grewe and John Nham and Nal Kalchbrenner and Ilya Sutskever and Timothy Lillicrap and Madeleine Leach and Koray Kavukcuoglu and Thore Graepel and Demis Hassabis},
  title   = {Mastering the Game of {Go} with Deep Neural Networks and Tree Search},
  journal = {Nature},
  number  = {7587},
  pages   = {484--489},
  volume  = {529},
  year    = {2016}
}

@article{silver-et-al-nature2017,
  author  = {David Silver and Julian Schrittwieser and Karen Simonyan and Ioannis Antonoglou and Aja Huang and Arthur Guez and Thomas Hubert and Lucas Baker and Matthew Lai and Adrian Bolton and Yutian Chen and Timothy Lillicrap and Fan Hui and Laurent Sifre and George van den Driessche and Thore Graepel and Demis Hassabis},
  title   = {Mastering the Game of {Go} Without Human Knowledge},
  journal = {Nature},
  number  = {7676},
  pages   = {354--359},
  volume  = {550},
  year    = {2017}
}

@article{silver-et-al-science2018,
  title   = {A general Reinforcement Learning algorithm that masters {Chess},
             {Shogi}, and {Go} through self-play},
  author  = {David Silver and Thomas Hubert and Julian Schrittwieser and
             Ioannis Antonoglou and Matthew Lai and Arthur Guez and
             Marc Lanctot and Laurent Sifre and Dharshan Kumaran and
             Thore Graepel and Timothy Lillicrap and Karen Simonyan and
             Demis Hassabis},
  journal = {Science},
  number  = {6419},
  pages   = {1140--1144},
  volume  = {362},
  year    = {2018}
}

@inproceedings{stahlberg-et-al-icaps2022,
  author    = {Simon St{\aa}hlberg and Blai Bonet and Hector Geffner},
  title     = {Learning General Optimal Policies with Graph Neural Networks: Expressive Power, Transparency, and Limits},
  booktitle = {Proceedings of the 32nd International Conference on Automated Planning and Scheduling ({ICAPS} 2022)},
  year      = {2022},
  pages     = {629--637}
}

@inproceedings{stahlberg-et-al-ipcl2023,
  author    = {Simon St{\aa}hlberg and Blai Bonet and Hector Geffner},
  title     = {{Muninn}},
  booktitle = {Learning Track of the {I}nternational {P}lanning {C}ompetition 2023: Planner Abstracts},
  year      = {2023}
}

@inproceedings{stahlberg-et-al-aaai2025,
  author    = {Simon St{\aa}hlberg and Blai Bonet and Hector Geffner},
  title     = {Learning More Expressive General Policies for Classical Planning Domains},
  booktitle = {Proceedings of the 39th {AAAI} Conference on Artificial Intelligence ({AAAI} 2025)},
  year      = {2025},
  pages     = {26697--26706}
}

@article{mnih-et-al-nature2015,
  title   = {Human-level control through deep Reinforcement Learning},
  author  = {Volodymyr Mnih and Koray Kavukcuoglu and David Silver and Andrei A. Rusu and Joel Veness and Marc G. Bellemare and Alex Graves and Martin A. Riedmiller and Andreas Kirkeby Fidjeland and Georg Ostrovski and Stig Petersen and Charlie Beattie and Amir Sadik and Ioannis Antonoglou and Helen King and Dharshan Kumaran and Daan Wierstra and Shane Legg and Demis Hassabis},
  journal = {Nature},
  year    = {2015},
  volume  = {518},
  pages   = {529-533}
}

@inproceedings{andrychowicz-et-al-nips2017,
  title     = {Hindsight Experience Replay},
  author    = {Andrychowicz, Marcin and Wolski, Filip and Ray, Alex and Schneider, Jonas and Fong, Rachel and Welinder, Peter and McGrew, Bob and Tobin, Josh and Abbeel, Pieter and Zaremba, Wojciech},
  booktitle = {Proceedings of the 31st Conference on Neural Information Processing Systems ({NIPS} 2017)},
  year      = {2017},
  volume    = {30},
  pages     = {5048--5058}
}

@inproceedings{aichmueller-geffner-ijcai2025,
  author    = {Michael Aichm{\"{u}}ller and Hector Geffner},
  title     = {Sketch Decompositions for Classical Planning via Deep Reinforcement Learning},
  booktitle = {Proceedings of the 34th International Joint Conference on Artificial Intelligence (IJCAI 2025)},
  year      = {2025},
  pages     = {8438--8446}
}

@inproceedings{stahlberg-geffner-aaai2026,
  author    = {Simon St{\aa}hlberg and Hector Geffner},
  title     = {First-Order Representation Languages for Goal-Conditioned {RL}},
  booktitle = {Proceedings of the 40th {AAAI} Conference on Artificial Intelligence ({AAAI} 2026)},
  year      = {2026}
}

@inproceedings{dillon:neurips24,
 author = {Chen, Dillon Z. and Thi\'{e}baux, Sylvie},
 booktitle = {Advances in Neural Information Processing Systems},
 doi = {10.52202/079017-2893},
 editor = {A. Globerson and L. Mackey and D. Belgrave and A. Fan and U. Paquet and J. Tomczak and C. Zhang},
 pages = {91156--91183},
 publisher = {Curran Associates, Inc.},
 title = {Graph Learning for Numeric Planning},
 volume = {37},
 year = {2024}
}

@article{horcik-et-al-aaai25, title={State Encodings for {GNN}-Based Lifted Planners}, volume={39},DOI={10.1609/aaai.v39i25.34853}, abstractNote={The application of graph neural networks (GNNs) to learn heuristic functions in classical planning is gaining traction. Despite the variety of methods proposed in the literature to encode classical planning tasks for GNNs, a comparative study evaluating their relative performances has been lacking. Moreover, some encodings have been assessed solely for their expressiveness rather than practical effectiveness in planning. This paper provides an extensive comparative analysis of existing encodings. Our results indicate that the smallest encoding based on Gaifman graphs, not yet applied in planning, outperforms the rest due to its fast evaluation times and the informativeness of the resulting heuristic. The overall coverage measured on the IPC almost reaches that of the state-of-the-art planner LAMA while exhibiting rather complementary strengths across different domains.}, number={25}, journal={Proceedings of the AAAI Conference on Artificial Intelligence}, author={Horčik, Rostislav and Šír, Gustav and Šimek, Vítězslav and Pevný, Tomáš}, year={2025}, month={Apr.}, pages={26525-26533} }

@article{kansky2023pushworld,
  title={{PushWorld}: A benchmark for manipulation planning with tools and movable obstacles},
  author={Kansky, Ken and Vaidyanath, Skanda and Swingle, Scott and Lou, Xinghua and L{\'a}zaro-Gredilla, Miguel and George, Dileep},
  journal={arXiv preprint arXiv:2301.10289},
  year={2023}
}

@article{horcik-et-al-icaps-2024, title={Expressiveness of Graph Neural Networks in Planning Domains}, volume={34}, url={https://ojs.aaai.org/index.php/ICAPS/article/view/31486}, DOI={10.1609/icaps.v34i1.31486}, abstractNote={Graph Neural Networks (GNNs) have become the standard method of choice for learning with structured data, demonstrating particular promise in classical planning. Their inherent invariance under symmetries of the input graphs endows them with superior generalization capabilities, compared to their symmetry-oblivious counterparts. However, this comes at the cost of limited expressive power. Particularly, GNNs cannot distinguish between graphs that satisfy identical sentences of C2 logic. To leverage GNNs for learning policies in PDDL domains, one needs to encode the contextual representation of the planning states as graphs. The expressiveness of this encoding, coupled with a specific GNN architecture, then hinges on the absence of indistinguishable states necessitating distinct actions. This paper provides a comprehensive theoretical and statistical exploration of such situations in PDDL domains across diverse natural encoding schemes and GNN models.}, number={1}, journal={Proceedings of the International Conference on Automated Planning and Scheduling}, author={Horčík, Rostislav and Šír, Gustav}, year={2024}, month={May}, pages={281-289} }

@article{bai-et-al-icaps2025, 
  title={Learning Efficiency Meets Symmetry Breaking}, 
  volume={35}, 
  number={1}, 
  journal={Proceedings of the International Conference on Automated Planning and Scheduling ({ICAPS} 2025)}, 
  author={Bai, Yingbin and Thiébaux, Sylvie and Trevizan, Felipe}, 
  year={2025}, 
  pages={154--159} 
}

@article{kirk2023survey,
  title={A survey of zero-shot generalisation in deep Reinforcement Learning},
  author={Kirk, Robert and Zhang, Amy and Grefenstette, Edward and Rockt{\"a}schel, Tim},
  journal={Journal of Artificial Intelligence Research},
  volume={76},
  pages={201--264},
  year={2023}
}

@article{lake2023human,
  title={Human-like systematic generalization through a meta-learning neural network},
  author={Lake, Brenden M and Baroni, Marco},
  journal={Nature},
  volume={623},
  number={7985},
  pages={115--121},
  year={2023},
  publisher={Nature Publishing Group UK London}
}

@article{mohan2024structure,
  title={Structure in deep Reinforcement Learning: A survey and open problems},
  author={Mohan, Aditya and Zhang, Amy and Lindauer, Marius},
  journal={Journal of Artificial Intelligence Research},
  volume={79},
  pages={1167--1236},
  year={2024}
}

@book{geffner:book,
  title={A concise introduction to models and methods for automated planning},
  author={Geffner, Hector and Bonet, Blai},
  year={2013},
  publisher={Morgan \& Claypool Publishers}
}

@book{russell:book,
  author    = {Stuart Russell and Peter Norvig},
  title     = {Artificial {I}ntelligence: A Modern Approach},
  publisher = {Pearson},
  notes = {4th Edition},
  year      = {2020}
}

@book{ghallab:book,
  title={Automated planning and acting},
  author={Ghallab, Malik and Nau, Dana and Traverso, Paolo},
  year={2016},
  publisher={Cambridge University Press}
}

@article{arora2018review,
  title={A review of learning planning action models},
  author={Arora, Ankuj and Fiorino, Humbert and Pellier, Damien and M{\'e}tivier, Marc and Pesty, Sylvie},
  journal={The Knowledge Engineering Review},
  volume={33},
  year={2018},
  publisher={Cambridge University Press}
}

@inproceedings{sift,
  title={Learning lifted {STRIPS} models from action traces alone: A simple, general, and scalable solution},
  author={G{\"o}sgens, Jonas and Jansen, Niklas and Geffner, Hector},
  booktitle={Proc. ICAPS},
  pages={189--197},
  year={2025}
}

@inproceedings{xi2024neuro,
  title={Neuro-Symbolic Learning of Lifted Action Models from Visual Traces},
  author={Xi, Kai and Gould, Stephen and Thi{\'e}baux, Sylvie},
  booktitle={Proceedings of the International Conference on Automated Planning and Scheduling},
  volume={34},
  pages={653--662},
  year={2024}
}

@article{sylvie:asnet,
  title={{ASNets}: Deep learning for generalised planning},
  author={Toyer, Sam and Thi{\'e}baux, Sylvie and Trevizan, Felipe and Xie, Lexing},
  journal={Journal of Artificial Intelligence Research (JAIR)},
  volume={68},
  pages={1--68},
  year={2020}
}

@article{erez:drl,
  title={Generalized planning with deep Reinforcement Learning},
  author={Rivlin, Or and Hazan, Tamir and Karpas, Erez},
  journal={arXiv preprint arXiv:2005.02305},
  year={2020}
}

@inproceedings{sid:generalization,
  title={Relational Abstractions for Generalized Reinforcement Learning on Symbolic Problems},
  booktitle={Proc. IJCAI},
  year = {2022},
  author={Karia, Rushang and Srivastava, Siddharth}
}

@inproceedings{simon:kr2023,
  title={Learning general policies with policy gradient methods},
  author={St{\aa}hlberg, Simon and Bonet, Blai and Geffner, Hector},
  booktitle={Proc. KR}, 
  pages={647--657},
  year={2023}
}

@inproceedings{dillon:h,
  title={Learning domain-independent heuristics for grounded and lifted planning},
  author={Chen, Dillon Z and Thi{\'e}baux, Sylvie and Trevizan, Felipe},
  booktitle={Proc.  AAAI},
  pages={20078--20086},
  year={2024}
}

@inproceedings{mausam,
  title={Symbolic network: generalized neural policies for relational {MDPs}},
  author={Garg, Sankalp and Bajpai, Aniket and others},
  booktitle={International Conference on Machine Learning},
  pages={3397--3407},
  year={2020},
  organization={PMLR}
}

@article{lrta,
  title={Real-time heuristic search},
  author={Korf, Richard E},
  journal={Artificial intelligence},
  volume={42},
  number={2-3},
  pages={189--211},
  year={1990}
}

@article{rtdp,
  title={Learning to act using real-time dynamic programming},
  author={Barto, Andrew G and Bradtke, Steven J and Singh, Satinder P},
  journal={Artificial intelligence},
  volume={72},
  number={1-2},
  pages={81--138},
  year={1995}
}

@book{sutton:book,
	address = {Cambridge, Massachusetts},
	edition = {Second},
	title = {Reinforcement learning: an introduction},
	publisher = {The MIT Press},
	author = {Sutton, Richard S. and Barto, Andrew G.},
	year = {2018}
}

@article{schrittwieser-et-al-nature2020,
  title={Mastering {ATARI}, {Go}, {Chess} and {Shogi} by planning with a learned model},
  author={Schrittwieser, Julian and Antonoglou, Ioannis and Hubert, Thomas and Simonyan, Karen and Sifre, Laurent and Schmitt, Simon and Guez, Arthur and Lockhart, Edward and Hassabis, Demis and Graepel, Thore and others},
  journal={Nature},
  volume={588},
  number={7839},
  pages={604--609},
  year={2020},
  publisher={Nature Publishing Group UK London}
}

@inproceedings{dillon-ecai2025,
  author    = {Dillon Z. Chen},
  title     = {{Weisfeiler-Leman} Features for Planning: A 1,000,000 Sample Size Hyperparameter Study},
  booktitle = {ECAI 2025},
  publisher = {IOS Press},
  year      = {2025},
  month     = oct,
  doi       = {10.3233/FAIA251370},
  url       = {https://doi.org/10.3233/FAIA251370}
}

@article{hadar-et-al-aaai2026, 
title={Beyond Single-Step Updates: Reinforcement Learning of Heuristics with Limited-Horizon Search}, volume={40}, 
url={https://ojs.aaai.org/index.php/AAAI/article/view/41023},
DOI={10.1609/aaai.v40i43.41023},
number={43},
journal={Proceedings of the AAAI Conference on Artificial Intelligence}, 
author={Hadar, Gal and Agostinelli, Forest and Shperberg, Shahaf S.},
year={2026}, 
month={Mar.}, 
pages={36955–36963} 
}

@article{general1, 
  title={A review of generalized planning}, 
  volume={34}, 
  DOI={10.1017/S0269888918000231}, 
  journal={The Knowledge Engineering Review}, author={Jiménez, Sergio and Segovia-Aguas, Javier and Jonsson, Anders}, 
  year={2019}, 
  pages={e5}
}

@inproceedings{explore1,
  title={Exploration by Random Network Distillation},
  author={Burda, Yuri and Edwards, Harrison and Storkey, Amos and Klimov, Oleg},
  booktitle={International Conference on Learning Representations (ICLR)},
  year={2019},
  url={https://openreview.net/forum?id=H1lJJnR5Ym},
}

@inproceedings{explore2,
  title={{NovelD}: A Simple yet Effective Exploration Criterion},
  author={Zhang, Tianjun and others},
  booktitle={Advances in Neural Information Processing Systems (NeurIPS)},
  year={2021},
}

@inproceedings{explore3,
  title={{RIDE}: Rewarding Impact‐Driven Exploration for Procedurally‐Generated Environments},
  author={Raileanu, Roberta and Rockt{\"a}schel, Tim},
  booktitle={International Conference on Learning Representations (ICLR)},
  year={2020},
  url={https://openreview.net/forum?id=rkg-TJBFPB}
}

@inproceedings{explore4,
  title={Exploration via Elliptical Episodic Bonuses},
  author={Henaff, Mikael and Raileanu, Roberta and Jiang, Minqi and Rockt{\"a}schel, Tim},
  booktitle={Advances in Neural Information Processing Systems (NeurIPS)},
  year={2022},
}

@inproceedings{nir:ecai2012,
  title={Width and serialization of classical planning problems},
  author={Geffner, Hector and Lipovetzky, Nir},
  booktitle={Proc. ECAI},
  year={2012},
  publisher={IOS Press}
}

@inproceedings{nir:2017,
  title={Best-first width search: Exploration and exploitation in classical planning},
  author={Lipovetzky, Nir and Geffner, Hector},
  booktitle={Proceedings of the AAAI Conference on Artificial Intelligence},
  year={2017}
}

@article{jendrik:width,
  title={Consolidating {LAMA} with Best-First Width Search},
  author={Corr{\^e}a, Augusto B and Seipp, Jendrik},
  journal={arXiv preprint arXiv:2404.17648},
  year={2024}
}

@article{sven:rts,
  title={Agent-centered search},
  author={Koenig, Sven},
  journal={AI Magazine},
  volume={22},
  number={4},
  pages={109--109},
  year={2001}
}

@inproceedings{goal-conditioned-rl,
  title={Universal Value Function Approximators},
  author={Schaul, Tom and Horgan, Dan and Gregor, Karol and Silver, David},
  booktitle={International Conference on Machine Learning (ICML)},
  year={2015},
  url={https://proceedings.mlr.press/v37/schaul15.html}
}

@article{bootstrap,
  title={Learning heuristic functions for large state spaces},
  author={Arfaee, Shahab Jabbari and Zilles, Sandra and Holte, Robert C},
  journal={Artificial Intelligence},
  volume={175},
  number={16-17},
  pages={2075--2098},
  year={2011}}

@book{pddl:book,
  title={An introduction to the planning domain definition language},
  author={Haslum, Patrik and Lipovetzky, Nir and Magazzeni, Daniele and Muise, Christian and Brachman, Ronald and Rossi, Francesca and Stone, Peter},
  volume={13},
  year={2019},
  publisher={Springer}
}

@article{sokoban:np-hard,
  title={Sokoban and other motion planning problems},
  author={Dor, Dorit and Zwick, Uri},
  journal={Computational Geometry},
  volume={13},
  number={4},
  pages={215--228},
  year={1999}
}
\bibliographystyle{icml2026}

\clearpage
\appendix
\section{Appendix}

This appendix supplements the main paper with additional material that expands on the presented results. First, we provide a pseudocode implementation of the algorithm introduced in \cref{sec:generalized_search_planning}, shown in \cref{alg:gsp}. Second, we include further details on the experimental results and training dynamics for the IPC-Learning domains.

\subsection{Detailed Validation Strategy}
Model selection is performed using a dedicated validation process. Validation is run concurrently during training, with each run saving its best checkpoint. To ensure robustness, each experiment is repeated across five random seeds, and the overall best-performing checkpoint is selected for reporting.

Within the concurrent validation loop, we periodically load the latest model parameters and evaluate greedy execution on the full validation set. Checkpoints are ranked primarily by validation coverage, with ties broken by the lowest total number of steps on solved instances, and any remaining ties resolved using the lowest RMSE on solved instances. For each experimental setting, we report the single run with the best validation score across all five seeds according to these criteria.

\subsection{Training Curves and Training Progress}
The IPC-Learning domains are particularly challenging, as successful test-time performance requires strong generalization beyond the training distribution. We summarize the required scaling behavior for these domains in \cref{tab:scaling}. Notably, our model successfully solves all \textsc{Blocks} instances, including the most difficult test case consisting of 488 blocks.

\begin{figure*}[t]
    \centering
    \begin{subfigure}{0.31\textwidth}
        \centering
        \includegraphics[width=\linewidth]{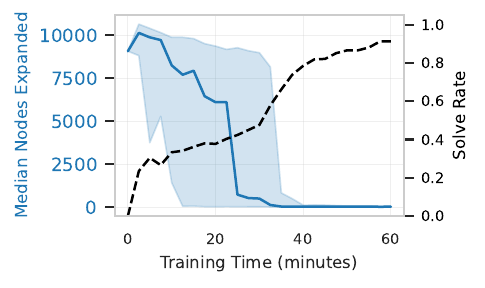}
        \caption{blocksworld}
        \label{fig:1}
    \end{subfigure}
    \hfill
    \begin{subfigure}{0.31\textwidth}
        \centering
        \includegraphics[width=\linewidth]{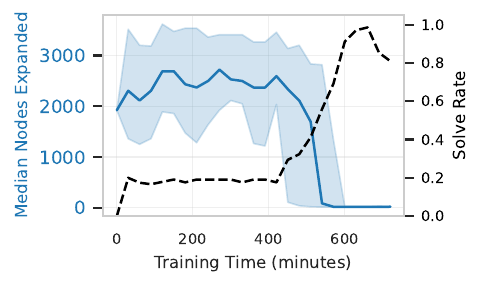}
        \caption{childsnack}
        \label{fig:2}
    \end{subfigure}
    \hfill
    \begin{subfigure}{0.31\textwidth}
        \centering
        \includegraphics[width=\linewidth]{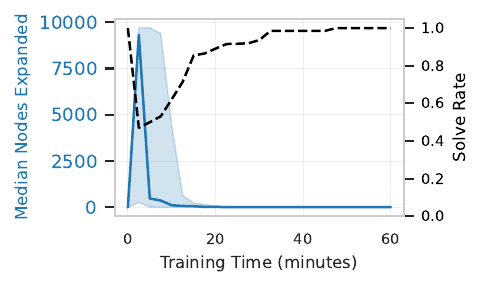}
        \caption{ferry}
        \label{fig:3}
    \end{subfigure}

    \vspace{0.3cm} 

    \begin{subfigure}{0.31\textwidth}
        \centering
        \includegraphics[width=\linewidth]{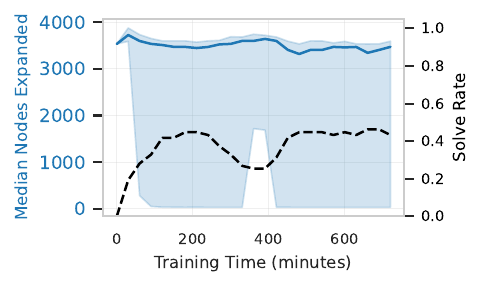}
        \caption{floortile}
        \label{fig:4}
    \end{subfigure}
    \hfill
    \begin{subfigure}{0.31\textwidth}
        \centering
        \includegraphics[width=\linewidth]{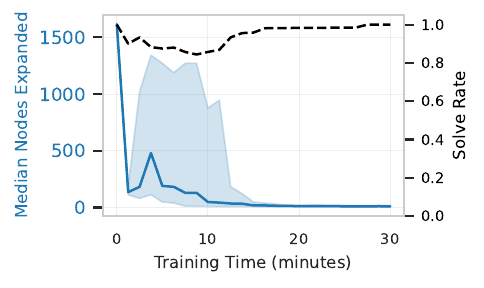}
        \caption{miconic}
        \label{fig:5}
    \end{subfigure}
    \hfill
    \begin{subfigure}{0.31\textwidth}
        \centering
        \includegraphics[width=\linewidth]{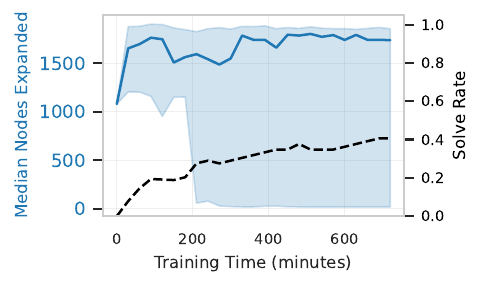}
        \caption{rovers}
        \label{fig:6}
    \end{subfigure}

    \vspace{0.3cm} 

    \begin{subfigure}{0.31\textwidth}
        \centering
        \includegraphics[width=\linewidth]{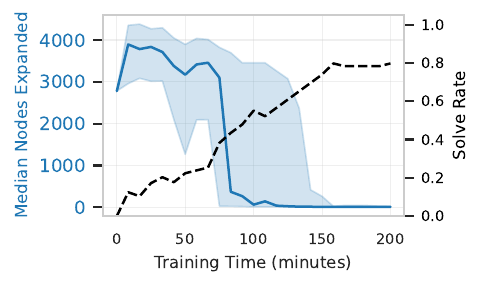}
        \caption{satellite}
        \label{fig:7}
    \end{subfigure}
    \hfill
    \begin{subfigure}{0.31\textwidth}
        \centering
        \includegraphics[width=\linewidth]{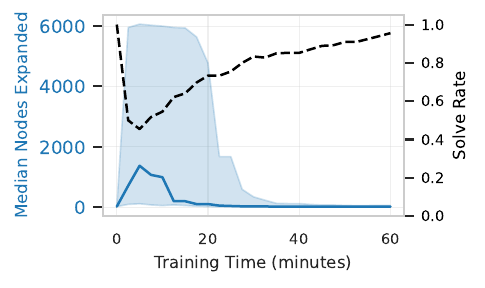}
        \caption{sokoban}
        \label{fig:8}
    \end{subfigure}
    \hfill
    \begin{subfigure}{0.31\textwidth}
        \centering
        \includegraphics[width=\linewidth]{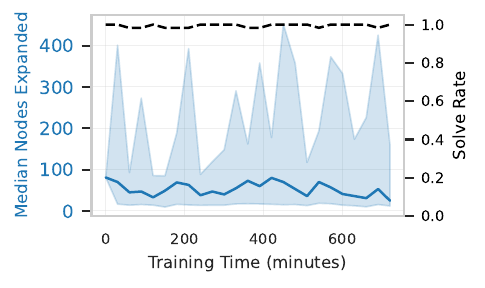}
        \caption{spanner}
        \label{fig:9}
    \end{subfigure}

    \vspace{0.3cm} 

    \begin{subfigure}{0.31\textwidth}
        \centering
        \includegraphics[width=\linewidth]{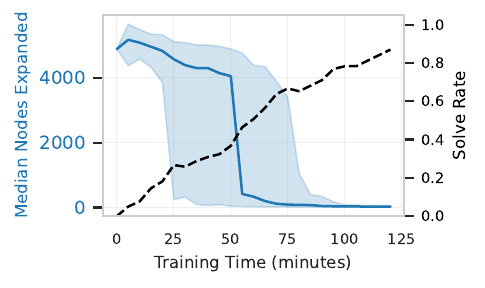}
        \caption{transport}
        \label{fig:10}
    \end{subfigure}

    \caption{Training progress across all IPC-learning domains. The blue line (left axis) shows the median number of expanded nodes, with shaded regions representing the 25th and 75th percentiles across all instances. The black line (right axis) shows the solve rate, indicating the percentage of problems solved at each point during training. Each training run lasted 12 hours (720 minutes), and we truncated the plots for runs considered converged.}

    \label{fig:all10}
\end{figure*}

We additionally present training curves for all IPC-Learning problems in \cref{fig:all10}. Most models converge on their respective training sets; however, we observe non-convergence for the \textsc{Floortile}, \textsc{Rovers}, and \textsc{Spanner} domains. This lack of convergence is generally reflected in their test-time performance, with the exception of \textsc{Spanner}. Although training in this domain is unstable--potentially due to forgetting or overfitting--the final model nevertheless exhibits strong generalization.

We hypothesize that this behavior is specific to the structure of the \textsc{Spanner} domain. In particular, there exists a simple but suboptimal greedy strategy that involves picking up every spanner encountered. While this strategy is not optimal, it generalizes remarkably well. In contrast, attempting to learn an optimal policy eliminates this greedy behavior, which may hinder generalization. This suggests a broader challenge for domains that admit highly generalizable greedy strategies but lack an optimal strategy with similar generalization properties. Our training procedure explicitly aims to converge to the optimal general policy, which may explain the observed instability.

\subsection{Additional Direct Comparison with Lifted HER}

A key baseline we consider is Lifted HER \cite{stahlberg-geffner-aaai2026}, as it employs the same architecture and also targets generalized planning. The main distinction lies in how sparse-reward learning is handled in such domains. Unfortunately, the original work \cite{stahlberg-geffner-aaai2026} did not evaluate on the learning track of the IPC. 

We applied our method to the domains presented in \cref{tab:lifted-her-benchmark-combined}. In many training runs, the fixed validation set is unable to assess the generalization capabilities of checkpoints accurately, often selecting a sub-par model. Thus, after all fixed validation instances had been evaluated, we extended the checkpoint-selection procedure with additional benchmark test instances to obtain a more stable model selection. These instances were used only for checkpoint selection and never for gradient updates. Overall, the results are comparable to Lifted HER, with a notable advantage in \textsc{Delivery}: the original authors reported that Lifted HER struggles due to the low probability of successfully delivering two boxes, a challenge that our method addresses effectively.

\subsection{Additional Batch Size Comparisons on Puzzle Domains}

The direct comparison of expanded nodes with \cite{orseau-lelis-aaai2021} in \cref{tab:puzzles-3way} is not entirely fair, because their search-based rollouts use a batch size of $32$. Instead of expanding a single node at a time, their method evaluates, and therefore expands, an entire batch of nodes simultaneously. This design better exploits GPU parallelism, but it also increases the reported expansion counts, since a batch size of $32$ imposes a minimum number of expansions per search step. To account for this difference, \cref{tab:puzzles-3way-batch-size} reports results for our models when evaluated with a batch size of $32$ at test time. As expected, the number of expanded nodes increases under the larger batch size. At the same time, we observe shorter solution plans on average. Overall, $\model{}$ continues to outperform the results reported in \cite{orseau-lelis-aaai2021} while still expanding fewer nodes.

\subsection{Fair comparison with Classical Heuristics}

LAMA \cite{richter-westphal-jair2010} is a strong domain-independent classical planner. However, a direct comparison with LAMA is difficult because performance is affected not only by the heuristic itself, but also by implementation and engineering choices, especially under fixed memory and time budgets. To isolate the heuristic contribution, \cref{tab:ipc-results-classic} compares $\model{}$ directly against the classical heuristics $h_{\mathrm{ff}}$ and $h_{\mathrm{max}}$ within the same execution framework. In all cases, we use $\wastar{}$ with $w=2$ and a maximum budget of $10{,}000$ node expansions.

The results show that $h_{\mathrm{ff}}$ achieves performance close to LAMA in several domains, whereas $h_{\mathrm{max}}$ performs substantially worse. In \textit{miconic}, \textit{rovers}, \textit{satellite}, and \textit{sokoban}, $h_{\mathrm{ff}}$ outperforms $\model{}$. These domains appear to benefit from a more exhaustive and consistent symbolic heuristic, especially because they contain many states that are nearly identical at the symbolic level. In such cases, $h_{\mathrm{ff}}$ provides more stable heuristic estimates than $\model{}$. Moreover, $h_{\mathrm{ff}}$ is often faster: when it fails, it typically reaches the limit of $10{,}000$ expansions, whereas $\model{}$ more often fails because of the one-hour timeout.

In the remaining domains, $\model{}$ outperforms the classical heuristics. These results suggest that $\model{}$ and $h_{\mathrm{ff}}$ capture complementary strengths. Combining the learned heuristic with $h_{\mathrm{ff}}$, for example through a multi-queue search strategy similar in spirit to LAMA, is therefore a promising direction for future work.

\begin{table*}[t]
  \centering
  \begin{tabular}{l|cc|ccc|ccc|ccc|}
  \toprule
  Domain
  & \multicolumn{2}{c|}{GSP$_\pi$}
  & \multicolumn{3}{c|}{GSP$_{\wastar{}}$}
  & \multicolumn{3}{c|}{$\wastar{}_{h_{\mathrm{ff}}}$}
  & \multicolumn{3}{c|}{$\wastar{}_{h_{\mathrm{max}}}$} \\
  &
  \multicolumn{1}{>{\columncolor{covbg}}c}{Cov.} & Steps
  & \multicolumn{1}{>{\columncolor{covbg}}c}{Cov.} & Exp. & Steps
  & \multicolumn{1}{>{\columncolor{covbg}}c}{Cov.} & Exp. & Steps
  & \multicolumn{1}{>{\columncolor{covbg}}c}{Cov.} & Exp. & Steps \\
  \midrule
  blocksworld
  & \textcolor{covgreen!100!covred}{100\%} & 444
  & \textcolor{covgreen!79!covred}{79\%} & 29 & 29
  & \textcolor{covgreen!12!covred}{12\%} & 2159 & 34
  & \textcolor{covgreen!3!covred}{3\%} & 702 & 13 \\
  childsnack
  & \textcolor{covgreen!41!covred}{41\%} & 31
  & \textcolor{covgreen!29!covred}{29\%} & - & -
  & \textcolor{covgreen!0!covred}{0\%} & - & -
  & \textcolor{covgreen!0!covred}{0\%} & - & - \\
  ferry
  & \textcolor{covgreen!87!covred}{87\%} & 422
  & \textcolor{covgreen!77!covred}{77\%} & 110 & 110
  & \textcolor{covgreen!67!covred}{67\%} & 480 & 115
  & \textcolor{covgreen!6!covred}{6\%} & 1076 & 11 \\
  floortile
  & \textcolor{covgreen!20!covred}{20\%} & 54
  & \textcolor{covgreen!28!covred}{28\%} & 42 & 42
  & \textcolor{covgreen!8!covred}{8\%} & 3599 & 43
  & \textcolor{covgreen!0!covred}{0\%} & - & - \\
  miconic
  & \textcolor{covgreen!100!covred}{100\%} & 490
  & \textcolor{covgreen!98!covred}{98\%} & 566 & 268
  & \textcolor{covgreen!100!covred}{100\%} & 840 & 288
  & \textcolor{covgreen!20!covred}{20\%} & 1802 & 12 \\
  rovers
  & \textcolor{covgreen!24!covred}{24\%} & 380
  & \textcolor{covgreen!11!covred}{11\%} & 245 & 17
  & \textcolor{covgreen!31!covred}{31\%} & 82 & 17
  & \textcolor{covgreen!6!covred}{6\%} & 2053 & 11 \\
  satellite
  & \textcolor{covgreen!61!covred}{61\%} & 134
  & \textcolor{covgreen!33!covred}{33\%} & 234 & 18
  & \textcolor{covgreen!67!covred}{67\%} & 24 & 18
  & \textcolor{covgreen!6!covred}{6\%} & 1248 & 7 \\
  sokoban
  & \textcolor{covgreen!14!covred}{14\%} & 18
  & \textcolor{covgreen!32!covred}{32\%} & 556 & 29
  & \textcolor{covgreen!34!covred}{34\%} & 194 & 31
  & \textcolor{covgreen!23!covred}{23\%} & 607 & 18 \\
  spanner
  & \textcolor{covgreen!100!covred}{100\%} & 216
  & \textcolor{covgreen!27!covred}{27\%} & 981 & 14
  & \textcolor{covgreen!33!covred}{33\%} & 64 & 12
  & \textcolor{covgreen!32!covred}{32\%} & 483 & 12 \\
  transport
  & \textcolor{covgreen!73!covred}{73\%} & 448
  & \textcolor{covgreen!57!covred}{57\%} & 29 & 28
  & \textcolor{covgreen!34!covred}{34\%} & 728 & 34
  & \textcolor{covgreen!7!covred}{7\%} & 1573 & 8 \\
  \bottomrule
  \end{tabular}
    \caption{Results on 2023 IPC benchmark. We report coverage (`Cov.' fraction of 90 test instances solved) and average plan length (Steps) over solved instances for $\model{}$ as a greedy policy ($\model{}_\pi$) and as a $\wastar{}(w=2)$ heuristic ($\model{}_{\wastar{}}$), alongside $h_{\mathrm{ff}}$ and $h_{\mathrm{max}}$. For $\model{}_{\wastar{}}$ and $\wastar{}_{h_{\mathrm{ff}}}$, expanded nodes (Exp.) and plan length (Steps) are reported only on problems solved by both $\model{}_{\wastar{}}$ and $\wastar{}_{h_{\mathrm{ff}}}$. For $\wastar{}_{h_{\mathrm{max}}}$, Exp.\ and Steps are reported on problems solved by both $\model{}_{\wastar{}}$ and $\wastar{}_{h_{\mathrm{max}}}$. The search budget (also for $h_{\mathrm{ff}}$ and $h_{\mathrm{max}}$) for each instance is limited to 10,000 expansions or one hour, whichever occurs first.}
    \label{tab:ipc-results-classic}
\end{table*}

\begin{table*}[t]
  \centering
  \begin{small}
  \begin{tabular}{l c c c c}
    \toprule
    Domain $\longrightarrow$ & \multicolumn{4}{c}{\textbf{\textcolor{sokcol}{Sokoban} / \textcolor{witcol}{The Witness}}} \\
    \midrule
    Model & Solved & Length & Expansions & Time (s) \\
    \midrule
    \midrule
    \textsc{GSP}$_{\mathrm{GBFS}}$
      & \sdouble{998}{1000}
      & \sdouble{38.5}{16.2}
      & \swdouble{564}{496}
      & \sdouble{59.7}{84.0} \\
    \textsc{GSP}$_{\wastar{}, w=2}$
      & \sdouble{1000}{1000}
      & \sdouble{36.0}{16.0}
      & \swdouble{207}{548}
      & \sdouble{22.1}{94.2} \\
    \textsc{GSP}$_{\mathrm{GBFS, b=32}}$
      & \sdouble{1000}{1000}
      & \sdouble{32.6}{14.7}
      & \swdouble{1028}{722}
      & \sdouble{103}{72.0} \\
    \textsc{GSP}$_{\wastar{}, w=2, b=32}$
      & \sdouble{1000}{1000}
      & \sdouble{32.5}{14.7}
      & \swdouble{972}{765}
      & \sdouble{61.1}{78.7} \\
    \midrule
    GBFS ($\dagger$)
      & \sdouble{914}{290}
      & \sdouble{37.7}{13.3}
      & \swdouble{5040}{10128}
      & \sdouble{49.2}{44.6} \\
    $\wastar{}, w=2$ ($\dagger$)
      & \sdouble{1000}{835}
      & \sdouble{35.6}{14.2}
      & \swdouble{3298}{14305}
      & \sdouble{22.8}{55.5} \\
    $\mathrm{PHS}^*$ ($\dagger$)
      & \sdouble{1000}{1000}
      & \sdouble{37.6}{14.4}
      & \swdouble{1522}{191}
      & \sdouble{11.3}{1.7} \\
    $\mathrm{LevinTS}$ ($\dagger$)
      & \sdouble{1000}{1000}
      & \sdouble{40.1}{14.8}
      & \swdouble{2640}{220}
      & \sdouble{19.5}{1.6} \\
    $\mathrm{PHS}_h$ ($\dagger$)
      & \sdouble{1000}{1000}
      & \sdouble{38.9}{14.6}
      & \swdouble{1962}{222}
      & \sdouble{14.8}{1.8} \\
    \bottomrule
  \end{tabular}
  \end{small}
  \caption{Batch-size-adjusted comparison on puzzle domains: Sokoban ($10\times 10$, 4 boxes) / The Witness ($5\times 5$). `Solved' is the number of solved instances out of 1000 per domain. Length, Expansions, and Time report averages over the solved instances in each domain. Sokoban models were trained on the same training problems as \citet{orseau-lelis-aaai2021}. All domains are converted to PDDL with relational encoding. The symbol $\dagger$ refers to \citet{orseau-lelis-aaai2021}. The $\model{}$ budget for each instance is limited to 100,000 expansions or five hours, whichever occurs first. For $\model{}_{b=32}$, we use a batch size of 32 to enable a fairer comparison against \citet{orseau-lelis-aaai2021}, whose results are also reported with a batch size of 32.}
  \label{tab:puzzles-3way-batch-size}
\end{table*}

\begin{algorithm}[t]
\caption{$\model{}$ search episode}
\label{alg:gsp}
\begin{algorithmic}[1]
\REQUIRE instance $\mathcal{E}$, initial state $s_0$, transition function $\mathcal{T}$, heuristic $Q_\theta$, weight $w$, budget $B$, dead-end bound $R_{\perp}$
\ENSURE replay tuples $\mathcal{D}$ of $(s,a,\underline{R})$

\STATE $\mathcal{F}\gets \mathrm{Queue}(\emptyset)$ \hfill
\STATE $\mathcal{V}\gets\{s_0\}$ \hfill \textit{\textcolor{gray!75!black}{(first-discovery visitation set)}}
\STATE $\mathcal{D}\gets\emptyset$
\FORALL{$a\in\mathcal{A}(s_0)$}
  \STATE $\textsc{parent}(s_0,a)\gets\bot$
  \STATE $g(s_0)\gets 0$
  \STATE push $(s_0,a)$ into $\mathcal{F}$ with priority $g(s_0)+wQ_\theta(s_0,a)$
\ENDFOR

\FOR{$t=1$ to $B$}
  \IF{$\mathcal{F}=\emptyset$} \STATE \textbf{break} \ENDIF
  \STATE pop $(s,a)$ with maximal priority from $\mathcal{F}$
  \STATE $s'\gets \mathcal{T}(s,a)$
  \IF{$s'$ is goal}
    \STATE $R\gets 0$ \hfill \textit{\textcolor{gray!75!black}{(suffix return along goal path)}}
    \WHILE{$(s,a)\neq\bot$}
      \STATE $R\gets R - 1$
      \IF{$(s,a,-\infty) \in \mathcal{D}$}
        \STATE remove $(s,a,-\infty)$ from $\mathcal{D}$
      \ENDIF
      \STATE add $(s,a,R)$ to $\mathcal{D}$
      \STATE $(s,a)\gets \textsc{parent}(s,a)$
    \ENDWHILE
    \STATE \textbf{return} $\mathcal{D}$
  \ENDIF
  \IF{$\mathcal{A}(s')=\emptyset$}
    \STATE add $(s,a, R_\perp)$ to $\mathcal{D}$; \textbf{continue}
  \ENDIF
  \STATE add $(s,a,-\infty)$ to $\mathcal{D}$ \hfill \textit{\textcolor{gray!75!black}{(unbounded bootstrap sample)}}
  \IF{$s'\notin\mathcal{V}$}
    \STATE $\mathcal{V}\gets \mathcal{V}\cup\{s'\}$;
    \FORALL{$a'\in\mathcal{A}(s')$}
      \STATE $\textsc{parent}(s',a')\gets(s,a)$
      \STATE $g(s')\gets g(s) - 1$
      \STATE push $(s',a')$ into $\mathcal{F}$ with priority $g(s') + w Q_\theta(s',a')$
    \ENDFOR
  \ENDIF
\ENDFOR
\STATE \textbf{return} $\mathcal{D}$
\end{algorithmic}
\end{algorithm}

\begin{table*}[t]
  \centering
  \begin{tabular}{l|cc|cc|cc|cc|cc}
  \toprule
  Domain
  & \multicolumn{2}{c|}{GSP$_\pi$}
  & \multicolumn{2}{c|}{GSP$_{\wastar{}}$}
  & \multicolumn{2}{c|}{Lifted HER}
  & \multicolumn{2}{c|}{Propositional HER}
  & \multicolumn{2}{c}{LAMA} \\
  &
  \multicolumn{1}{>{\columncolor{covbg}}c}{Cov.} & Steps
  & \multicolumn{1}{>{\columncolor{covbg}}c}{Cov.} & Steps
  & \multicolumn{1}{>{\columncolor{covbg}}c}{Cov.} & Steps
  & \multicolumn{1}{>{\columncolor{covbg}}c}{Cov.} & Steps \\
  \midrule
  blocks
  & \textcolor{covgreen!100!covred}{100\%} & 97.9
  & \textcolor{covgreen!100!covred}{100\%} & 96.5
  & \textcolor{covgreen!100!covred}{100\%} & 96.7
  & \textcolor{covgreen!100!covred}{100\%} & 107.3
  & \textcolor{covgreen!100!covred}{100\%} & 232.5 \\
  childsnack
  & \textcolor{covgreen!79!covred}{79\%} & 89.3
  & \textcolor{covgreen!34!covred}{34\%} &59.4
  & \textcolor{covgreen!56!covred}{56\%} & 77.8
  & \textcolor{covgreen!100!covred}{100\%} & 92.6
  & \textcolor{covgreen!100!covred}{100\%} & 98.4 \\
  delivery
  & \textcolor{covgreen!93!covred}{93\%} & 293.3
  & \textcolor{covgreen!42!covred}{42\%} & 134.7
  & \textcolor{covgreen!12!covred}{12\%} & 292.2
  & \textcolor{covgreen!11!covred}{11\%} & 167.0
  & \textcolor{covgreen!99!covred}{99\%} & 276.1 \\
  gripper
  & \textcolor{covgreen!100!covred}{100\%} & 238.0
  & \textcolor{covgreen!74!covred}{74\%} & 202.6
  & \textcolor{covgreen!100!covred}{100\%} & 238.0
  & \textcolor{covgreen!100!covred}{100\%} & 239.0
  & \textcolor{covgreen!100!covred}{100\%} & 238.0 \\
  miconic
  & \textcolor{covgreen!100!covred}{100\%} & 158.0
  & \textcolor{covgreen!100!covred}{100\%} & 158.0
  & \textcolor{covgreen!100!covred}{100\%} & 160.0
  & \textcolor{covgreen!100!covred}{100\%} & 158.4
  & \textcolor{covgreen!100!covred}{100\%} & 195.6 \\
  reward
  & \textcolor{covgreen!75!covred}{75\%} & 119.3
  & \textcolor{covgreen!99!covred}{99\%} & 94.2
  & \textcolor{covgreen!58!covred}{58\%} & 95.9
  & \textcolor{covgreen!72!covred}{72\%} & 91.6
  & \textcolor{covgreen!99!covred}{99\%} & 121.8 \\
  spanner
  & \textcolor{covgreen!95!covred}{95\%} & 95.3
  & \textcolor{covgreen!0!covred}{0\%} & 0.0
  & \textcolor{covgreen!100!covred}{100\%} & 95.5
  & \textcolor{covgreen!100!covred}{100\%} & 95.5
  & \textcolor{covgreen!0!covred}{0\%} & 0.0 \\
  visitall
  & \textcolor{covgreen!94!covred}{94\%} & 562.4
  & \textcolor{covgreen!41!covred}{41\%} & 331.1
  & \textcolor{covgreen!100!covred}{100\%} & 453.3
  & \textcolor{covgreen!88!covred}{88\%} & 392.6
  & \textcolor{covgreen!100!covred}{100\%} & 530.0 \\
  \bottomrule
  \end{tabular}
  \caption{Benchmark results on the same data as \cite{stahlberg-et-al-aaai2025}. We report coverage (Cov.) and average plan length (Steps) for $\model{}$ as a greedy policy ($\model{}_\pi$) and as a $\wastar{}(w=2)$ heuristic ($\model{}_{\wastar{}}$), compared against \textsc{Lifted HER} and \textsc{Propositional HER} \cite{stahlberg-et-al-aaai2025} and \textsc{LAMA}.}
  \label{tab:lifted-her-benchmark-combined}
\end{table*}

\begin{table*}[t]
  \centering
  \begin{tabular}{l|cc|cc|cc|cc}
\toprule
$\model{}_\pi$\rule[-0.7ex]{0pt}{3.0ex}
& \multicolumn{2}{c|}{\emph{Complete}}
& \multicolumn{2}{c|}{\emph{No dead-end supervision}}
& \multicolumn{2}{c|}{\emph{No solution bounds}}
& \multicolumn{2}{c}{\emph{No sampling strategy}} \\[0.5ex]
\cline{1-1}
Domain\rule{0pt}{2.5ex}
& \multicolumn{1}{>{\columncolor{covbg}}c}{Cov.} & \multicolumn{1}{c|}{Steps}
& \multicolumn{1}{>{\columncolor{covbg}}c}{Cov.} & \multicolumn{1}{c|}{Steps}
& \multicolumn{1}{>{\columncolor{covbg}}c}{Cov.} & \multicolumn{1}{c|}{Steps}
& \multicolumn{1}{>{\columncolor{covbg}}c}{Cov.} & \multicolumn{1}{c}{Steps} \\
\midrule
  blocksworld & \textcolor{covgreen!83!covred}{83\% $\pm$ 24\%} & 410 & \textcolor{covgreen!89!covred}{89\% $\pm$ 16\%} & 368 & \textcolor{covgreen!82!covred}{82\% $\pm$ 24\%} & 343 & \textcolor{covgreen!63!covred}{63\% $\pm$ 16\%} & 284 \\
  childsnack & \textcolor{covgreen!33!covred}{33\% $\pm$ 15\%} & 71 & \textcolor{covgreen!53!covred}{53\% $\pm$ 12\%} & 279 & \textcolor{covgreen!44!covred}{44\% $\pm$ 10\%} & 38 & \textcolor{covgreen!33!covred}{33\% $\pm$ 16\%} & 36 \\
  ferry & \textcolor{covgreen!92!covred}{92\% $\pm$ 9\%} & 605 & \textcolor{covgreen!83!covred}{83\% $\pm$ 11\%} & 435 & \textcolor{covgreen!72!covred}{72\% $\pm$ 17\%} & 328 & \textcolor{covgreen!89!covred}{89\% $\pm$ 19\%} & 801 \\
  floortile & \textcolor{covgreen!19!covred}{19\% $\pm$ 5\%} & 62 & \textcolor{covgreen!29!covred}{29\% $\pm$ 6\%} & 93 & \textcolor{covgreen!19!covred}{19\% $\pm$ 8\%} & 66 & \textcolor{covgreen!33!covred}{33\% $\pm$ 6\%} & 134 \\
  miconic & \textcolor{covgreen!96!covred}{96\% $\pm$ 9\%} & 309 & \textcolor{covgreen!84!covred}{84\% $\pm$ 15\%} & 195 & \textcolor{covgreen!74!covred}{74\% $\pm$ 25\%} & 193 & \textcolor{covgreen!95!covred}{95\% $\pm$ 9\%} & 251 \\
  rovers & \textcolor{covgreen!26!covred}{26\% $\pm$ 4\%} & 491 & \textcolor{covgreen!24!covred}{24\% $\pm$ 5\%} & 475 & \textcolor{covgreen!31!covred}{31\% $\pm$ 4\%} & 716 & \textcolor{covgreen!26!covred}{26\% $\pm$ 2\%} & 692 \\
  satellite & \textcolor{covgreen!44!covred}{44\% $\pm$ 10\%} & 331 & \textcolor{covgreen!53!covred}{53\% $\pm$ 6\%} & 241 & \textcolor{covgreen!47!covred}{47\% $\pm$ 9\%} & 308 & \textcolor{covgreen!48!covred}{48\% $\pm$ 5\%} & 276 \\
  sokoban & \textcolor{covgreen!16!covred}{16\% $\pm$ 3\%} & 19 & \textcolor{covgreen!15!covred}{15\% $\pm$ 1\%} & 20 & \textcolor{covgreen!19!covred}{19\% $\pm$ 2\%} & 20 & \textcolor{covgreen!17!covred}{17\% $\pm$ 2\%} & 23 \\
  spanner & \textcolor{covgreen!97!covred}{97\% $\pm$ 4\%} & 190 & \textcolor{covgreen!72!covred}{72\% $\pm$ 21\%} & 186 & \textcolor{covgreen!36!covred}{36\% $\pm$ 5\%} & 22 & \textcolor{covgreen!55!covred}{55\% $\pm$ 29\%} & 99 \\
  transport & \textcolor{covgreen!85!covred}{85\% $\pm$ 15\%} & 407 & \textcolor{covgreen!76!covred}{76\% $\pm$ 11\%} & 542 & \textcolor{covgreen!73!covred}{73\% $\pm$ 10\%} & 184 & \textcolor{covgreen!80!covred}{80\% $\pm$ 7\%} & 424 \\
  \bottomrule

  \toprule
$\model{}_{\wastar{}}$\rule[-0.7ex]{0pt}{3.0ex}
& \multicolumn{2}{c|}{\emph{Complete}}
& \multicolumn{2}{c|}{\emph{No dead-end supervision}}
& \multicolumn{2}{c|}{\emph{No solution bounds}}
& \multicolumn{2}{c}{\emph{No sampling strategy}} \\[0.5ex]
\cline{1-1}
Domain\rule{0pt}{2.5ex}
& \multicolumn{1}{>{\columncolor{covbg}}c}{Cov.} & \multicolumn{1}{c|}{Steps}
& \multicolumn{1}{>{\columncolor{covbg}}c}{Cov.} & \multicolumn{1}{c|}{Steps}
& \multicolumn{1}{>{\columncolor{covbg}}c}{Cov.} & \multicolumn{1}{c|}{Steps}
& \multicolumn{1}{>{\columncolor{covbg}}c}{Cov.} & \multicolumn{1}{c}{Steps} \\
\midrule
  blocksworld & \textcolor{covgreen!72!covred}{72\% $\pm$ 4\%} & 196 & \textcolor{covgreen!78!covred}{78\% $\pm$ 5\%} & 236 & \textcolor{covgreen!78!covred}{78\% $\pm$ 7\%} & 235 & \textcolor{covgreen!71!covred}{71\% $\pm$ 5\%} & 188 \\
  childsnack & \textcolor{covgreen!23!covred}{23\% $\pm$ 14\%} & 23 & \textcolor{covgreen!27!covred}{27\% $\pm$ 15\%} & 24 & \textcolor{covgreen!29!covred}{29\% $\pm$ 9\%} & 22 & \textcolor{covgreen!27!covred}{27\% $\pm$ 15\%} & 24 \\
  ferry & \textcolor{covgreen!67!covred}{67\% $\pm$ 0\%} & 110 & \textcolor{covgreen!67!covred}{67\% $\pm$ 0\%} & 110 & \textcolor{covgreen!66!covred}{66\% $\pm$ 5\%} & 111 & \textcolor{covgreen!68!covred}{68\% $\pm$ 2\%} & 119 \\
  floortile & \textcolor{covgreen!32!covred}{32\% $\pm$ 2\%} & 59 & \textcolor{covgreen!34!covred}{34\% $\pm$ 2\%} & 66 & \textcolor{covgreen!26!covred}{26\% $\pm$ 9\%} & 60 & \textcolor{covgreen!35!covred}{35\% $\pm$ 2\%} & 72 \\
  miconic & \textcolor{covgreen!81!covred}{81\% $\pm$ 9\%} & 141 & \textcolor{covgreen!84!covred}{84\% $\pm$ 11\%} & 167 & \textcolor{covgreen!81!covred}{81\% $\pm$ 17\%} & 160 & \textcolor{covgreen!89!covred}{89\% $\pm$ 11\%} & 199 \\
  rovers & \textcolor{covgreen!8!covred}{8\% $\pm$ 3\%} & 20 & \textcolor{covgreen!12!covred}{12\% $\pm$ 4\%} & 19 & \textcolor{covgreen!11!covred}{11\% $\pm$ 4\%} & 18 & \textcolor{covgreen!11!covred}{11\% $\pm$ 3\%} & 19 \\
  satellite & \textcolor{covgreen!27!covred}{27\% $\pm$ 9\%} & 15 & \textcolor{covgreen!33!covred}{33\% $\pm$ 0\%} & 18 & \textcolor{covgreen!29!covred}{29\% $\pm$ 6\%} & 15 & \textcolor{covgreen!33!covred}{33\% $\pm$ 1\%} & 17 \\
  sokoban & \textcolor{covgreen!33!covred}{33\% $\pm$ 1\%} & 30 & \textcolor{covgreen!32!covred}{32\% $\pm$ 0\%} & 30.0 & \textcolor{covgreen!33!covred}{33\% $\pm$ 1\%} & 29 & \textcolor{covgreen!33!covred}{33\% $\pm$ 1\%} & 30 \\
  spanner & \textcolor{covgreen!26!covred}{26\% $\pm$ 2\%} & 14 & \textcolor{covgreen!30!covred}{30\% $\pm$ 4\%} & 14 & \textcolor{covgreen!25!covred}{25\% $\pm$ 2\%} & 12 & \textcolor{covgreen!31!covred}{31\% $\pm$ 2\%} & 14 \\
  transport & \textcolor{covgreen!56!covred}{56\% $\pm$ 6\%} & 51 & \textcolor{covgreen!50!covred}{50\% $\pm$ 2\%} & 44 & \textcolor{covgreen!46!covred}{46\% $\pm$ 2\%} & 35 & \textcolor{covgreen!50!covred}{50\% $\pm$ 3\%} & 41 \\
  \bottomrule
  \end{tabular}
  \caption{IPC-2023 test results for $\model{}_{\pi}$ and $\model{}_{\wastar{}}$ across training variants. \emph{Cov.} denotes the fraction of the 90 test instances solved, reported as mean $\pm$ standard deviation over independently trained models. \emph{Steps} denotes the average plan length over solved instances, averaged across models. $\model{}_{\wastar{}}$ evaluates the learned model with $\wastar{}$ search using $w=2$.}
  \label{tab:ipc-2023-ablations-gsp}
\end{table*}

\end{document}